\documentclass{article}

\usepackage[margin=1in]{geometry}
\usepackage{times}

\usepackage[utf8]{inputenc}
\usepackage[T1]{fontenc}
\usepackage{microtype}
\usepackage{graphicx}
\usepackage{booktabs}
\usepackage{multirow}
\usepackage{amsmath,amssymb}
\usepackage{xcolor}
\usepackage{url}
\usepackage[colorlinks=true,linkcolor=blue,citecolor=blue,urlcolor=blue]{hyperref}
\usepackage{natbib}
\usepackage{enumitem}
\setlist{nosep,leftmargin=1.4em}

\newcommand{\method}[1]{\textsc{#1}}

\title{UpgradeBench: A Decision-Centric Benchmark\\for Upgrading Fine-Tuned LLM Specialists}

\author{%
\begin{tabular}{c@{\hspace{2.2cm}}c}
Ye Chen & Weining Zhang$^{*}$ \\[3pt]
\normalsize Alibaba Group &
\normalsize Cheung Kong Graduate School of Business
\end{tabular}}
\date{}

\begin{document}
\maketitle
\renewcommand{\thefootnote}{\fnsymbol{footnote}}
\footnotetext[1]{Corresponding author.}
\renewcommand{\thefootnote}{\arabic{footnote}}

% =====================================================================
\begin{abstract}
Organizations increasingly maintain task-specific adapters on
open-weight language models, and every base-model release re-opens the
same migration decision: keep the existing specialist, port its
adapter, refresh it from retained behavior, or retrain. Existing
transfer studies evaluate isolated model pairs; none measures this
decision over real release sequences. We introduce
\textbf{UpgradeBench}, a decision-centric longitudinal benchmark over
four consecutive Qwen releases, one continuation release, six
application tasks, and two model scales, complemented by OLMo
checkpoints whose training relationships are documented rather than
inferred. The benchmark separates three questions: whether the target
checkpoint raises a fixed-recipe retrained-specialist score, whether
the specialization asset is portable, and which recovery resources
remain available. We find that the value of upgrading varies by
task--scale--release episode: some retrained references improve while
others stay flat within training noise, and calendar-time durability
spans from
an H50 under one release interval (text-to-SQL) to beyond fourteen
months (intent classification). Raw adapter copying is governed
neither by architecture nor by family: on documented OLMo edges its
retention falls from 0.88--0.99 at a 46B-token continuation to zero at
a 2.9T-token continuation, with no additional damage from annealing or
model souping---portability decays with continued-pretraining
distance. When retained inputs are available, teacher relabeling
recovers target-base specialists with no new gold annotation, though
not always less compute. Replaying a pre-specified decision policy
over 33 measured upgrade episodes---with decision gates and scoring
on disjoint halves of each test set---attains a mean quality regret of
0.37pp with zero behavioral regressions at one third of the
always-retrain compute and label budget, and a cheap representational
probe (CKA over 256 prompts) ranks portability across base pairs
(Spearman 0.74 over eight pairs). We release per-example
predictions, cost logs, split manifests, and the evaluation harness.
\end{abstract}

% =====================================================================
\section{Introduction}
\label{sec:intro}

Two facts about deploying language models in organizations are now
well established. First, \emph{specialization works}: on narrow,
recurrent tasks---intent triage, text-to-SQL, function
calling---parameter-efficiently fine-tuned small models match or beat
prompted frontier models at one to two orders of magnitude lower
inference cost \citep{zhao2024loraland,bucher2024finetuned,
belcak2025slm}. A single 24\,GB GPU can host a base model plus
hundreds of task adapters \citep{sheng2024slora}, making a ``fleet of
specialists'' an economically attractive alternative to routing all
traffic to a frontier API. Second, \emph{base models improve fast}:
the capability density of open-weight models doubles on the order of
months \citep{xiao2024densing}; each generation of the Qwen, Llama, or
Gemma families reaches the previous generation's quality with a
fraction of the parameters \citep{qwen3,gemma3}.

These two facts collide, because a fine-tuned adapter is bound, both
numerically and behaviorally, to the base checkpoint it was trained on. When a
better base ships, the organization faces a decision it currently
cannot price: (i)~\textbf{freeze}---keep serving the old specialist;
(ii)~\textbf{port}---transfer the adapter or task vector to the new
base; (iii)~\textbf{refresh}---use the old specialist as a teacher to
cheaply re-specialize the new base; or (iv)~\textbf{retrain}---pay the
full data and compute cost again. We call the quality-and-cost gap
between the chosen strategy and the best attainable one the
\emph{base-model upgrade tax}.

The research community has recognized the problem but not measured it.
A rapidly growing family of methods transfers fine-tuning across
bases: synthetic-data distillation \citep{wang2024translora,
jung2026titok,guan2026tuneshift}, upgrade-specific parameter mappings
\citep{li2025lorasuite,portllm2025}, subspace projection
\citep{lora-x2025,xia2025crosslora}, diff-vector porting
\citep{lin2025finetuningtransfer,huang2024chatvector}, and
training-time immunization \citep{gu2025transpeft}. Each is evaluated
on its own choice of model pairs and tasks; comparisons across papers
are impossible. A recurring assumption behind these methods is that adapter weights
copied unchanged between sufficiently similar bases form a strong
trivial baseline; that assumption is what our shape-compatible hops
test directly.
Meanwhile the most decision-relevant quantity has never appeared in
print: \emph{how many generations does it take for a new base's cheap
adoption to overtake yesterday's specialist?} Position papers argue
both sides from first principles \citep{belcak2025slm}; neither side
has controlled evidence.

\textbf{This paper} supplies the missing measurement. UpgradeBench
tracks a complete, real release series of four Qwen generations
spanning fourteen months, rather than ad-hoc model pairs, and evaluates
the full freeze/port/refresh/retrain decision space on six tasks from
three classes, at two model scales, with training-side costs
(GPU-minutes, energy) as first-class outputs.

Three common intuitions do not survive the measurement. The first is
that fine-tunes obsolesce quickly: across nine upgrade hops, freezing
retains 99--101\% of the attainable gain on intent classification, and
over four generations the retrained reference score moves 0.2 points while
zero-shot rises twenty-one (Figure~\ref{fig:halflife}a). But
specialization moats are not uniformly safe either: on
text-to-SQL the frozen specialist forfeits up to 59\% of attainable
gain on a single hop, and its advantage over a newer base decays to zero
within three generations (Figure~\ref{fig:halflife}b). Nor is naive weight porting governed by the rule practitioners assume:
between two architecturally identical checkpoints from independent
pretraining runs, pasting the adapter onto the newer base collapses
intent classification from 92.8\% to 42.9\%, far below the 60.7\%
obtainable with no adapter at all; yet onto a continued-pretraining
descendant of the same weights, the same operation retains
reference-level accuracy. What licenses copying is continuity of the
underlying weights, not the shape of the network
(\S\ref{sec:results:copy}). A prospectively specified validation on OLMo, whose
intermediate checkpoints make the training relationship a manipulable
variable rather than an inference from release notes, reproduces both
regimes in a second family and then locates the boundary between them:
moving the target along the same documented trajectory, retention
falls from 0.88--0.99 at a 46B-token continuation to the no-adapter
floor at a 2.9T-token continuation, with the subsequent anneal and
model-soup merge adding no further damage. Portability is not a
family property but a distance budget. The organizing variable behind the first
two results is whether task competence is \emph{base-bound} (SQL
reasoning) or \emph{data-bound} (label taxonomies, call formats). This cuts across the
discriminative/generative surface distinction that transfer
evaluations usually adopt.

\textbf{Contributions.}
\begin{itemize}
\item \textbf{C1 (Benchmark).} A longitudinal, cost-accounted benchmark
of fine-tuning transfer along a complete open-weight release
lineage, at two model scales, with unified tasks, metrics, per-query
prediction records, bootstrap CIs, and energy logs, all released. The
entire study (98 trainings, 193 evaluation cells) runs on one RTX~4090.
\item \textbf{C2 (Quantities).} Formal definitions and first
measurements of retention, recovery cost, and specialization
half-life, including per-hop retention across nine real hops, a
half-life bracketed to two-to-three generations at 7--8B, and one
crossing observed outright at 1.5--1.8B scale.
\item \textbf{C3 (Strategy map).} A measured
freeze/port/refresh/retrain decision map, including a continuity
criterion for porting: copying fails between independent pretraining
runs (eleven task/scale cells, two families) and succeeds between
checkpoints connected by continued pretraining (five cells, two
families), with a prospectively specified OLMo distance ablation showing that
portability decays with continued-pretraining distance (near-parity at
46B tokens, floor-level at 2.9T) while annealing and model souping add
no further damage,
while annotation-free (input-retaining) refresh
reaches retraining parity everywhere it was tested.
\item \textbf{C4 (Weights vs.\ harness).} A first controlled
comparison of weight-side and prompt-side specialization under the
same upgrade (appendix pilot): the weight asset is roughly
80$\times$ larger in delivered advantage on the one task/hop measured,
and no more fragile than the prompt asset, whose optimal choice flips
across generations.
\end{itemize}

% =====================================================================
\section{Related Work}
\label{sec:related}

\textbf{Model updates and backward compatibility.} Replacing a
deployed model regresses individual predictions even when aggregate
accuracy improves; negative-flip rate quantifies this
\citep{yan2021positive}, and update-regression mitigation has been
studied for structured NLP \citep{cai2022measuring}. This literature
updates a model inside one training regime; we measure the same
phenomenon where the \emph{base} changes underneath a specialist, and
fold negative flips into the upgrade decision itself
(\S\ref{sec:results:flips}). The maintenance framing follows the
technical-debt view of ML systems \citep{sculley2015hidden}.
\label{sec:related}

\textbf{Cross-base adapter transfer.}
Trans-LoRA \citep{wang2024translora} first framed base deprecation as
a problem, proposing near-data-free transfer via synthetic-data
distillation; TiTok \citep{jung2026titok} and TuneShift-KD
\citep{guan2026tuneshift} refine the distillation route. LoRASuite
\citep{li2025lorasuite} and PortLLM \citep{portllm2025} target version
upgrades with parameter mappings and training-free patches; LoRA-X
\citep{lora-x2025} and Cross-LoRA \citep{xia2025crosslora} project
adapters through aligned subspaces; Trans-PEFT \citep{gu2025transpeft}
immunizes adapters at training time. Each evaluates on self-selected
pairs. Prior re-evaluation argues that naive weight copying is a strong
baseline between similar bases; our measurement on
architecturally identical consecutive releases contradicts that for
data-bound tasks (\S\ref{sec:results:copy}). UpgradeBench proposes no
new transfer method; it provides the missing common yardstick along a
real release sequence, and documents that two of three hops admit no
weight-space method at all.

\textbf{Task vectors, merging, and re-use.}
Task arithmetic treats fine-tuning deltas as portable vectors
\citep{ilharco2023task}, with theory tied to shared initialization
\citep{ortiz2023tangent}; Chat Vector \citep{huang2024chatvector} and
fine-tuning transfer \citep{lin2025finetuningtransfer} demonstrate
near-free porting within a lineage; merging combines specialists on a
shared base \citep{yadav2023ties,yu2024dare}, with recent skepticism
about cross-task recycling \citep{liu2026recycling}. All presuppose
shape compatibility, which holds for exactly one of the three
major-version hops in our lineage, and even there, transfer is harmful
on every task we test. Our paired hops turn the shared-initialization
assumption of the theory \citep{ortiz2023tangent} into an experimental
variable: transfer survives a continued-pretraining hop and dies across
independent pretraining runs of identical architecture, exactly as the
tangent-space account predicts.

\textbf{Capability growth of small open models.}
Capability density doubles on the order of months
\citep{xiao2024densing}, making the upgrade decision recur quarterly;
this trend supplies our half-life denominator. Our measurements add a
caveat: the trend is an average, and per-task uplift can be negative
(\S\ref{sec:results:nonmono}).

\textbf{Economics of specialist fleets.}
Fine-tuned small specialists beat prompted frontier models on narrow
tasks \citep{zhao2024loraland,bucher2024finetuned}; multi-adapter
serving makes fleets cheap to host \citep{sheng2024slora}; position
work argues small models are the natural unit of agentic workloads
\citep{belcak2025slm}. Cost-aware evaluation
\citep{chen2023frugalgpt,ong2025routellm,erol2025costofpass,
kapoor2024aiagents} established dollar-denominated reporting; we adopt
it and add training-side accounting. The upgrade tax is the missing
depreciation term in these analyses.

\textbf{Prompt-side upgrade tax.}
Model updates silently break prompt-engineered applications
\citep{chen2023chatgptdrift,ma2023promptregression,
tripathi2025promptmigration}. We measure both taxes in one controlled
frame for the first time (\S\ref{sec:results:prompt}).

\textbf{Benchmarks.}
Enterprise and agent benchmarks evaluate systems at a fixed time
\citep{xu2024theagentcompany}; cost-aware methodology exists
\citep{kapoor2024aiagents,reuel2024betterbench}, but no benchmark
versions its \emph{models} along release time. UpgradeBench is, to our
knowledge, the first whose primary axis is the release sequence
itself.

% =====================================================================
\section{Formalizing the Upgrade Tax}
\label{sec:formal}

\textbf{Setup.} A \emph{lineage} is a sequence of base checkpoints
$\mathcal{B} = (b_0, \dots, b_G)$ ordered by public release date. For
a task $t$ with training set $D_t$ and metric $S_t(\cdot) \in [0,1]$,
a fixed PEFT procedure $A(b, D)$ (QLoRA; \S\ref{sec:setup}) produces a
specialist $s_{b,t}$. $S_t(b)$ denotes the base's \emph{cheap
adoption} score: zero-shot with full task instructions (plus $k{=}5$
exemplars where the format benefits, reported separately).

\textbf{Specialization gain.} $G(b, t) = S_t(s_{b,t}) - S_t(b)$.

\textbf{Retention.} For a hop $b_i \to b_j$ and strategy $m$ producing
$m(s_{b_i,t}, b_j)$ without full-budget access to $D_t$'s labels:
\begin{equation}
R_{i\to j}^{m}(t) \;=\;
\frac{S_t\!\big(m(s_{b_i,t},\, b_j)\big) - S_t(b_j)}
     {S_t\!\big(s_{b_j,t}\big) - S_t(b_j)} .
\label{eq:retention}
\end{equation}
$R{=}1$ matches full retraining on the new base; $R{=}0$ adds nothing
over cheap adoption; $R{<}0$ is \emph{worse than not transferring at
all}; $R{>}1$ beats retraining (both observed in our data). The
\emph{upgrade tax} of $m$ is
$T^{m} = (1-R^{m})\cdot(S_t(s_{b_j,t}) - S_t(b_j))$, paired with $m$'s
measured cost. \method{Freeze} is the special case
$m(s_{b_i,t}, b_j) = s_{b_i,t}$. Two reporting rules follow from the
algebra. First, $T^m$ simplifies to the \emph{signed quality
shortfall} $L^m = O - Q^m$ against the reference $O$; we report $L^m$
with a paired bootstrap CI as the primary quantity, since it is what a
deployment loses in points. Second, the normalized retention $R^m$ is
unstable when the reference gain $O - F$ is small, so we show $R^m$
only where the lower confidence bound of $O-F$ exceeds 2pp and mark
the cell denominator-unstable otherwise. The full upgrade tax of an
episode is a vector---shortfall, migration GPU-minutes and energy,
gold labels, teacher queries, and validation load---and we keep it a
vector: any scalarization needs organization-specific utility weights,
so we report Pareto components and let the decision model
(\S\ref{sec:cost}) combine them.

\textbf{Expert advantage and durability.} The advantage of a
generation-$i$ specialist over a generation-$(i{+}g)$ base at calendar
time $\tau(i{+}g)$ (the target's actual release date) is
$A_t(\tau) = S_t(s_{b_i,t}) - S_t(b_{i+g})$. We report two horizons:
the \emph{half-value horizon}
$H_{50} = \inf\{\tau : A_t(\tau) \le \tfrac{1}{2} A_t(0)\}$ and the
\emph{crossover horizon} $H_{0} = \inf\{\tau : A_t(\tau) \le 0\}$,
both in months of real release time. Because releases are discrete,
each horizon is interval-censored: we report the interval between the
last release where the criterion fails and the first where it holds
(or a lower bound when it never holds within the observed series).
Earlier drafts called the crossover a ``half-life''; the two horizons
differ by an order of magnitude on some tasks, so we keep them
separate. For non-monotone trajectories we report the first crossing
and note persistence.

\textbf{Recovery cost.} $C^{m}_{95}(t)$ is the labeled-example (and
GPU-minute) budget of tuning on $b_j$ needed to reach 95\% of the
\emph{attainable reference gain} over the adoption floor, i.e.\ the
smallest budget $N$ with $Q_N \ge F + 0.95\,(O - F)$ where $F$ is the
target's floor and $O$ its full-data reference, measured on a ladder
$\{0, 256, 1024, 4096, \text{full}\}$. (An absolute-score criterion
$Q_N \ge 0.95\,O$ gives different, systematically smaller budgets; we
use the gain criterion because it is invariant to the floor.)

\textbf{Amortized decision model.} Given upgrade cadence $\Delta$,
request volume $v$, per-strategy quality $Q^m$ and cost $c^m$, the
steady-state objective is $\max_m v \cdot u(Q^m) - c^m/\Delta$;
released as code with all measured $(Q^m, c^m)$.

% =====================================================================
\section{The UpgradeBench Benchmark}
\label{sec:benchmark}

\subsection{Lineage}
\label{sec:lineages}

We instantiate the benchmark on the Qwen 7--8B release series \citep{qwen2025qwen25}---the
open-weight family most used for enterprise fine-tuning---covering
every major generation released between February~2024 and April~2025
(Table~\ref{tab:lineages}). We fine-tune \emph{instruct} checkpoints,
the standard enterprise substrate, and use the same checkpoints for
cheap-adoption baselines, keeping the substrate constant across
conditions. Qwen3's thinking mode is disabled in both training and
evaluation templates.

\begin{table}[t]
\centering
\small
\caption{The measured lineage, from the released checkpoints'
\texttt{config.json}. Shape compatibility governs whether
weight-space transfer (adapter copying, diff-vector porting) is
\emph{defined at all}: it requires identical hidden width, depth, and
attention geometry. Exactly one of the three hops qualifies.}
\label{tab:lineages}
\begin{tabular}{llrrrrrl}
\toprule
Checkpoint & Released & Hidden & Layers & Heads (Q/KV) & FFN & Vocab & Hop to next \\
\midrule
Qwen1.5-7B-Chat      & 2024-02 & 4096 & 32 & 32 / 32 (MHA) & 11008 & 151936 & \textbf{incompatible} \\
Qwen2-7B-Instruct    & 2024-06 & 3584 & 28 & 28 / 4 (GQA)  & 18944 & 152064 & \textbf{compatible} \\
Qwen2.5-7B-Instruct  & 2024-09 & 3584 & 28 & 28 / 4 (GQA)  & 18944 & 152064 & \textbf{incompatible} \\
Qwen3-8B             & 2025-04 & 4096 & 36 & 32 / 8 (GQA)  & 12288 & 151936 & --- \\
\midrule
\multicolumn{8}{l}{\emph{Continuation release}: Qwen2.5-7B-Instruct-1M
(2025-01) --- identical to Qwen2.5-7B-Instruct in every} \\
\multicolumn{8}{l}{dimension above, and a continued-pretraining
descendant of the same weights (long-context extension).} \\
\multicolumn{8}{l}{\emph{Small-scale track}: Qwen1.5-1.8B $\to$
Qwen2-1.5B $\to$ Qwen2.5-1.5B $\to$ Qwen3-1.7B; the 2$\to$2.5 hop is} \\
\multicolumn{8}{l}{likewise shape-identical (hidden 1536, 28 layers,
GQA 12/2, FFN 8960).} \\
\bottomrule
\end{tabular}
\end{table}

Two properties of this lineage matter methodologically. First, it is
\emph{tokenizer-stable in kind but not in size}: all generations use
the same BPE family, but the vocabulary alternates between 151{,}936
and 152{,}064 entries, so even embedding-level surgery is not uniform
across hops. Second, and more consequentially, \emph{architecture
churn is the norm rather than the exception}: hidden width, depth, and
attention geometry all change at the 1.5$\to$2 and 2.5$\to$3
boundaries, leaving weight-space transfer methods
\citep{lin2025finetuningtransfer,ilharco2023task}
undefined on two of the three hops an enterprise following this family
would have faced. The 2$\to$2.5 hop is the exception: those two
checkpoints are architecturally identical in every dimension above, so
an adapter trained on one can be loaded onto the other without any
mapping. That hop is therefore the best case for naive porting between
independent pretraining runs, and we measure it
(\S\ref{sec:results:copy}). The lineage also offers the complementary
case: Qwen2.5-7B-Instruct-1M, a January-2025 long-context release that
is a continued-pretraining \emph{descendant} of Qwen2.5-7B-Instruct
with identical shapes. Copying onto it isolates weight-space
continuity from architectural compatibility, the distinction that
shared-initialization theory \citep{ilharco2023task,ortiz2023tangent}
and within-family porting results
\citep{huang2024chatvector,lin2025finetuningtransfer} implicitly rely
on. A parallel 1.5--1.8B track replicates the whole design at small
scale. Applicability is itself a
benchmark outcome: transfer papers that evaluate only on
shape-compatible pairs describe a regime that two thirds of this
lineage's real upgrades never entered.

\paragraph{Cross-family validation lineage (OLMo).} Qwen's training
history is not public, so ``independent pretraining run'' and
``continued-pretraining descendant'' are inferences from release
documentation. To test the continuity criterion where lineage is
\emph{ground truth}, a fourth experimental wave uses OLMo
\citep{groeneveld2024olmo,olmo2025olmo2}, whose checkpoints, data
order, and training stages are fully documented. It contributes five
edges, all between architecturally identical checkpoints:
(i)~OLMo-1-7B $\to$ OLMo-1.7-7B, whose model card states the latter
was trained from scratch---a verified fresh-pretraining pair; and,
holding the source fixed at the OLMo-2-7B stage-1 checkpoint
\texttt{step237000}, four targets progressively farther along the same
documented run \citep{olmo2025olmo2}: (ii)~\texttt{step248000},
${\sim}$46B tokens later---a short pure continuation;
(iii)~\texttt{step928646}, the end of stage 1, ${\sim}$2.9T tokens
later with no annealing or merging---a long pure continuation;
(iv)~the final \texttt{stage2-ingredient3} checkpoint, after the 50B
stage-2 anneal on a different data mix but before souping; and
(v)~the released OLMo-2-7B base, a model-soup merge of independently
annealed branches. Edges (iii)--(v) form a distance-matched ablation
that separates continuation distance from the release operations. These are base (non-instruct)
checkpoints, so the OLMo wave fine-tunes and evaluates with a fixed
plain-text prompt format on the two intent tasks, with per-checkpoint
adoption floors and retraining references measured under the identical
format (\S\ref{sec:results:copy}). All three predictions were
pre-registered before any wave-4 run (\S\ref{sec:rq}).

\subsection{Task suite}
\label{sec:tasks}

Two tasks per class (Table~\ref{tab:tasks}); splits fixed at seed 42
and released. The original three tasks are measured across all four
generations and both scales; the second task in each class (CLINC150,
FinQA, glaive-FC) across the three most recent 7--8B generations.

\begin{table}[t]
\centering
\small
\caption{Task suite. D = discriminative, S = structured generation,
A = agentic tool use. Licenses as listed by official distributions.}
\label{tab:tasks}
\begin{tabular}{lllll}
\toprule
Task & Class & Train / Val / Test & Metric & Source \\
\midrule
Banking77 & D (77-way intent) & 9{,}003 / 1{,}000 / 3{,}080 & accuracy & \citet{casanueva2020banking77} \\
CLINC150 & D (151-way intent, incl.\ OOS) & 10{,}000 / 1{,}000 / 5{,}500 & accuracy & \citet{larson2019clinc} \\
Spider & S (text-to-SQL) & 6{,}500 / 500 / 1{,}034 (dev) & execution acc. & \citet{yu2018spider} \\
FinQA & S (numerical programs) & 6{,}251 / 500 / 1{,}147 & program acc. & \citet{chen2021finqa} \\
xLAM-FC & A (function calling) & 10{,}000 / 500 / 2{,}000 (held-out) & call-set match & \citet{liu2024apigen} \\
glaive-FC & A (calling + refusal) & 7{,}580 / 505 / 2{,}000 (group re-split) & call-set match & glaive-v2 corpus \\
\bottomrule
\end{tabular}
\end{table}

\textbf{Protocol notes.} (i)~Banking77 cheap adoption carries the
77-label inventory in the system prompt; zero-shot and 5-shot (fixed
exemplars) are both reported. A manual audit found \emph{zero} parsing
failures across all baseline conditions. Every baseline error is a
genuine near-class confusion, so adoption floors are not deflated by
format artifacts. (ii)~Spider prompts serialize the schema (tables,
columns, types, primary/foreign keys); execution accuracy compares
result multisets on the official dev databases (single-database
variant; known false-positive risk noted). (iii)~xLAM-FC uses a
held-out split of the APIGen-verified corpus \citep{liu2024apigen}
with semantic call-set matching (name equality, type-coerced argument
equality, multiset matching); it measures in-distribution function
calling. (iv)~A memorization probe over all four bases and three test
sets (Appendix~\ref{app:contamination}) finds no evidence of verbatim
test-set memorization on Banking77 or Spider (verbatim-continuation
rates 0--1\%); xLAM shows moderate $n$-gram overlap (0.32--0.46)
attributable to its templated JSON structure, roughly constant across
generations---so cross-generation comparisons are not differentially
biased. Any residual contamination inflates \emph{baselines}, making
measured expert advantages conservative. (v)~For the second-task
suite: CLINC150 label matching is reported under a strict
normalization and a lenient morphological one (e.g., mapping
\emph{translation} to the gold label \emph{translate} when the match
is unique); we use the lenient score as the adoption floor so that
baselines are not penalized for surface form, and release both. FinQA
is scored by exact match of the normalized calculation program.
glaive-FC retains 25\% no-call negatives, so specialists are also
graded on refusing to call.

\subsection{Upgrade strategies}
\label{sec:strategies}

\begin{table}[t]
\centering
\small
\caption{Strategy conditions and measurement status. For LoRA
specialists, pasting the adapter and adding its merged
$\Delta W = BA$ to the new base are the same operation, so the
\method{Copy} measurement covers diff-vector porting as well.}
\label{tab:strategies}
\begin{tabular}{lllp{4.6cm}p{2.6cm}}
\toprule
ID & Strategy & Data need & Description & Measured \\
\midrule
\method{Freeze} & keep old specialist & --- & serve $s_{b_i,t}$ unchanged & 9 hops \\
\method{Adopt-0} & new base, zero-shot & --- & prompt-only cheap adoption & 4 gens \\
\method{Adopt-k} & new base, 5-shot & --- & fixed exemplars & 4 gens (D) \\
\method{Copy}/\method{PortDiff} & weight-space port & none & load $s_{b_i,t}$'s LoRA on $b_j$ \citep{lin2025finetuningtransfer} & 2$\to$2.5: 6 tasks (+3 at 1.8B); 2.5$\to$1M: 3 tasks \\
\method{Refresh-D} & distillation refresh & unlabeled inputs & old specialist labels retained inputs; fresh LoRA on $b_j$ (relabel variant of \citealp{wang2024translora}) & 2.5$\to$3, 3 tasks \\
\method{Retrain} & full QLoRA on $b_j$ & full labels & reference denominator of Eq.~\ref{eq:retention} & all generations, 6 tasks, 2 scales \\
\method{Retrain}-$N$ & label-budget ladder & $N$ labels & $N \in \{256, 1024, 4096\}$ & Spider, Qwen3 \\
\method{PromptPort} & port best prompt & dev slice & old best-of-6 prompt on $b_j$ & Banking77 hop \\
\method{PromptRetune} & re-search prompt & dev slice & new best-of-6 search on $b_j$ & Banking77 hop \\
\bottomrule
\end{tabular}
\end{table}

\subsection{Cost accounting and protocol}
\label{sec:cost}

Every training run logs wall-clock, peak VRAM, and energy (board power
sampled at 10\,s intervals, integrated). Greedy decoding throughout;
per-query outputs for all 125 evaluation cells are released with
bootstrap 95\% CIs (10{,}000 resamples). Because every pair of
conditions is evaluated on the identical test set, all comparisons
between conditions are \emph{paired}; we therefore test them with the
exact McNemar test on discordant pairs rather than by inspecting
overlapping marginal CIs, and report $p$-values with the point
estimates. We call a difference measured only when $p < 0.05$. A 3-seed probe on
the Banking77/Qwen2.5 cell bounds run-to-run variance at $\sigma
\approx 0.12$pp; a second seed on the Spider/Qwen3 reference reveals much
larger training variance on the generative task (2.5pp), which we fold
into how the single-hop Spider numbers are read
(\S\ref{sec:results:threats}). Training drops examples whose completion is fully
truncated at the context limit; the counts vary with each model's
tokenizer (Spider 138/6{,}500 for most generations but 154 for Qwen3;
xLAM 160 vs.\ 161; FinQA 6--8; glaive-FC and Banking77 0) and are
reported per run in the released logs rather than folded silently into
the totals.

% =====================================================================
\section{Experimental Setup}
\label{sec:setup}

One recipe for all 55 trainings: QLoRA \citep{dettmers2023qlora} with
NF4 double quantization, bf16 compute; LoRA $r{=}16$, $\alpha{=}32$,
dropout 0.05 on all attention/MLP projections; lr $2{\times}10^{-4}$
cosine, warmup 3\%, completion-only loss, seed 42 (43/44 for variance
probes). Intent tasks: 3 epochs, length 256, batch 16. Spider/FinQA: 3
epochs, length 1024/1280, micro-batch 4 $\times$ accumulation 4.
Function calling: 2 epochs, length 1024, same batching (small-scale
track: micro-batch 8 $\times$ 2). Optimizer
\texttt{paged\_adamw\_8bit}; gradient checkpointing on.

\textbf{Stack and hardware.} Single RTX~4090 (24\,GB); PyTorch
2.13.0+cu126, transformers 5.14.1, peft 0.19.1, bitsandbytes 0.50.0.
Peak VRAM 6.5--21.1\,GB across runs.

\textbf{Study inventory.} 98 QLoRA trainings (12 lineage specialists
at 7--8B, 9 second-task specialists, 12 small-scale specialists, 3
continuation-release references, 4 Refresh-D students, 3 ladder
budgets, 17 OLMo specialists/references along the documented
trajectory and release series, 3 glaive re-split specialists, 4
rank-64 replicates, and 31 seed replicates), 4
relabeling passes, 193 evaluation cells, one memorization probe (4
bases $\times$ 3 test sets), one prompt-side pilot. Cost is reported
as a three-way split, not a single number: 148.7 training GPU-hours
with 47.8\,kWh of measured board energy, 79.4 evaluation GPU-hours
(evaluation energy was not separately metered), and 4.5 teacher-relabel
GPU-hours (Appendix~\ref{app:costlog}); \$0 API spend. Table~\ref{tab:coverage} states the exact coverage of every action--edge pair; the matrix is sparse by design and the empty cells are part of the record.

\begin{table}[t]
\centering
\small
\caption{Coverage matrix: which actions are measured on which edges.
The design is deliberately sparse, not factorial; empty cells are
stated, not implied. $^a$Label ladder run on the highest-coupling task
only. $^b$\method{Refresh-D} requires retained inputs; measured on the
newest hop of the original three tasks. $^c$Learned heterogeneous
mappings and distillation-transfer baselines are extension points of
the released protocol, not measured here.}
\label{tab:coverage}
\setlength{\tabcolsep}{2.5pt}\scriptsize
\begin{tabular}{lccccc}
\toprule
Panel & Adopt/Freeze & \method{Copy} & \method{Refresh-D} & Retrain-$N$ & Reference \\
\midrule
B77/Spider/xLAM, 7--8B & 4 gens + 1M & 2$\to$2.5, 2.5$\to$1M & 2.5$\to$3 (3 seeds) & Spider only$^a$ & 4 gens + 1M (3 seeds) \\
CLINC/FinQA/glaive, 7--8B & 3 gens & 2$\to$2.5 & ---$^b$ & --- & 3 gens \\
B77/Spider/xLAM, 1.5--1.8B & 4 gens & s2$\to$s2.5 & --- & --- & 4 gens \\
B77/CLINC, OLMo & 7 checkpoints & 5 documented edges & --- & --- & 7 checkpoints \\
Shape-breaking hops & (= Adopt/Freeze) & undefined$^c$ & --- & --- & (= Reference) \\
\bottomrule
\end{tabular}
\end{table}

% =====================================================================
\section{Prospectively Specified Analyses}
\label{sec:rq}

Hypotheses were fixed before the wave that tests them; outcomes are
reported against them in \S\ref{sec:results}.

A methodological caveat first: these analyses were fixed in the
runner scripts before the corresponding waves ran, and the scripts are
released verbatim with timestamps, but they were not deposited with an
external immutable registry. We therefore call them \emph{prospectively
specified} rather than pre-registered, and we mark each hypothesis
with the wave boundary it preceded.

\begin{itemize}
\item \textbf{RQ1 (Durability).} How large is $A_t(\tau)$, and where
are $H_{50}$ and $H_0$? \emph{H1 (before wave 1):} crossover within
one generation for reasoning-heavy structured tasks, two or more for
discriminative tasks. \emph{Outcome: directionally supported for the
discriminative task ($H_0 > 14$ months, advantage 27.1pp at the last
observed release, $p<10^{-200}$) but refuted in magnitude for
text-to-SQL: the crossover took the full observed series
($H_0 = 14$ months, interval-censored at $(7,14]$), while the
half-value horizon was under one release interval
($H_{50} \in (0,4]$ months---the advantage drops from 17.2pp to 4.0pp
at the first hop). The two horizons differ by an order of magnitude,
which the original single ``half-life'' number conflated.}
\item \textbf{RQ2 (Strategies).} What retention at what cost?
\emph{H2 (before wave 1):} on shape-compatible hops
\method{Copy}/\method{PortDiff} retain $R \ge 0.8$; across breaks only
\method{Refresh} exceeds $R = 0.5$. \emph{Outcome: both clauses
refuted as stated. Between independent pretraining runs \method{Copy}
retained $-0.60$ to $0.78$ across nine task/scale cells, below
\method{Freeze} in all nine; between checkpoints joined by a short
continuation it retained 0.82--1.45. Shape was the wrong variable, and
(after wave 5) continuity alone is not sufficient either: portability
decays with continued-pretraining distance (\S\ref{sec:results:copy}).
\method{Refresh-D} reached 0.96--1.05, far above 0.5, at zero gold
labels.}
\item \textbf{RQ3 (Task class).} Does a
discriminative-vs-generative retention gap replicate on real release
series? \emph{H3 (before wave 1):} retention
ranks D $>$ S $>$ A. \emph{Outcome: refuted. Reference scores rise
only where the coupling between task competence and base capability is
high, and \method{Copy}'s ordering (A $>$ S $>$ D) inverts the prior
report. We describe this coupling with the continuous elasticity
$\beta_t = \Delta O_t / \Delta F_t$ (reference gain per point of floor
gain) rather than a binary base-bound/data-bound label, since the
coupling itself shifts with model scale
(\S\ref{sec:results:tax}, \S\ref{sec:results:smallscale}).}
\item \textbf{RQ4 (Prompt asset).} Is prompt-side specialization more
durable than weight-side? \emph{Outcome (pilot, appendix only): the
prompt asset is roughly 80$\times$ smaller in delivered gain on the
one task/hop measured, and its optimum flips across generations; the
pilot lacks per-example records, so we report it as a bounded
observation, not a contribution.}
\item \textbf{RQ5 (Predictability).} Do cheap representational probes
predict retention before any migration is paid for? \emph{Outcome
(wave 6, exploratory): across the eight measured base pairs, linear
CKA over 256 task prompts rank-correlates with mean $R(\method{Copy})$
at Spearman $\rho{=}0.74$; next-token top-1 agreement and JSD are
weaker (0.38, $-0.52$). No single probe is decision-grade at
$n{=}8$---CKA separates the OLMo distance spectrum cleanly but cannot
distinguish the Qwen fresh pair from the Qwen continuation pair
(\S\ref{sec:results:probe}).}
\end{itemize}

\textbf{Wave-4 prospective predictions (fixed 2026-07-31, before any
OLMo run).} \emph{P1}: copying OLMo-1 adapters onto OLMo-1.7 (verified
fresh pretraining) collapses, $R < 0.5$ and below \method{Freeze};
\emph{P2}: copying between OLMo-2 stage-1 checkpoints of the same run
(${\sim}46$B tokens apart) retains $R \approx 1$, indistinguishable
from the reference; \emph{P3}: copying across the stage-2 anneal and
soup merge is a stress test whose either outcome delimits the
criterion. \emph{Outcomes (\S\ref{sec:results:copy}): P1 confirmed
($R{=}-0.32$ and $-0.02$, at or below the adoption floor). P2
directionally confirmed ($R{=}0.88$ and $0.99$, far above the floor)
but exact reference parity is rejected on Banking77 ($-11.3$pp,
$p{<}10^{-79}$): even 46B continuation tokens exact a small, measurable
tax. P3's collapse was confirmed, but its attribution was not
resolvable from that wave alone.}

\textbf{Wave-5 prospective predictions (fixed 2026-08-10, before any
wave-5 run).} \emph{Q1}: if the Qwen copy collapse is not a
chat-template artifact, swapping source/target templates changes
nothing; \emph{Q2}: constrained decoding shrinks but does not
eliminate the CLINC generational decline; \emph{Q3}: if the wave-4 P3
collapse is driven by annealing/souping, copying onto the
end-of-stage-1 checkpoint (2.9T tokens of \emph{pure} continuation, no
anneal, no soup) stays high; if driven by distance, it collapses
there already. \emph{Outcomes: Q1 confirmed exactly (template swap
changes no score by even one item). Q2 confirmed (the decline shrinks
by roughly a third under constrained decoding). Q3 resolved
\textbf{against} the annealing/souping reading our wave-4 text had
favored: copying is already at the floor after 2.9T tokens of pure
continuation, and the anneal and soup edges add no further damage
(\S\ref{sec:results:copy}). We report the correction rather than the
earlier interpretation.}

\textbf{Wave-6 prospective predictions (fixed 2026-08-16).}
\emph{W1}: on a second family (the OLMo release series), the
high-coupling task's half-value horizon is again shorter than one
release interval; \emph{W2}: next-token top-1 agreement predicts
$R(\method{Copy})$ better than CKA; \emph{W3}: the fresh-hop copy
collapse at rank 64 matches the rank-16 magnitude.
\emph{Outcomes: W1 refuted, informatively: on OLMo the frozen Spider
specialist keeps a 39--58pp advantage across nine months of releases,
because that series' floors and reference scores barely move on SQL
(\S\ref{sec:results:halflife}); durability horizons are properties
of the task--series pair, not of the task. W2
refuted---CKA is the stronger rank-predictor ($\rho{=}0.74$ vs.\\
0.38). W3 refuted in magnitude: at $r{=}64$ the Banking77 copy loses
21pp instead of 50pp ($R$: $-0.55 \to +0.29$) and Spider reaches
$R{=}0.74$, though copying stays below both \method{Freeze} and the
reference at both ranks (\S\ref{sec:results:copy}); the collapse
magnitude is rank-dependent, its ordering is not.}

% =====================================================================
\section{Results}
\label{sec:results}

\begin{figure}[t]
\centering
\includegraphics[width=\linewidth]{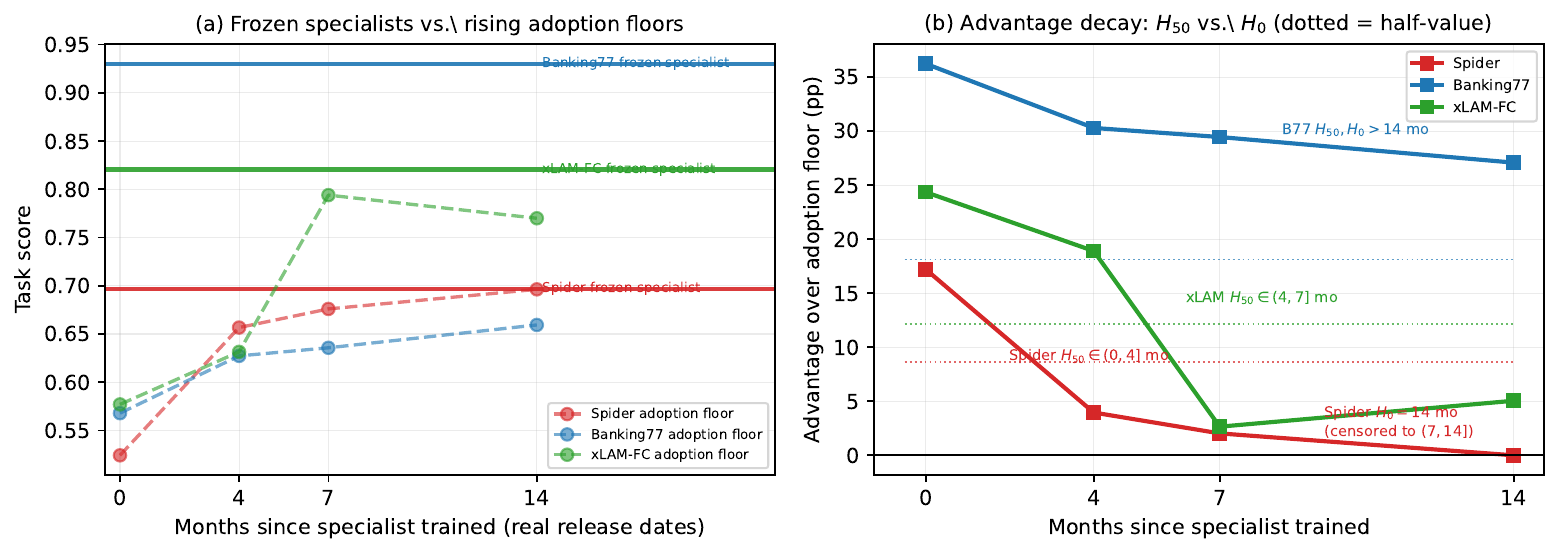}
\caption{Durability in real release time (x-axis: months since the
gen-0 specialists were trained; all points measured, $n$ = 3{,}080 /
1{,}034 / 2{,}000 test items). \textbf{(a)}~Frozen gen-0 specialists
(horizontal lines) against each task's rising adoption floor.
\textbf{(b)}~The same data as advantage over the floor, with
half-value ($H_{50}$, dotted) and crossover ($H_0$) horizons marked;
horizons are interval-censored by the discrete release calendar.
Generated from the released study registry.}
\label{fig:halflife}
\end{figure}

\begin{table}[t]
\centering
\small
\caption{Lineage scores (test sets; greedy decoding; 4-bit inference).
``Spec.''\ = specialist fine-tuned on that generation (the retrain
reference for hops ending there). Bootstrap 95\% CIs for all cells are
released; typical half-widths $\pm$0.9pp (Banking77), $\pm$2.8pp
(Spider), $\pm$1.6pp (xLAM).}
\label{tab:lineage}
\begin{tabular}{lcccccccc}
\toprule
 & \multicolumn{2}{c}{Qwen1.5-7B} & \multicolumn{2}{c}{Qwen2-7B} &
\multicolumn{2}{c}{Qwen2.5-7B} & \multicolumn{2}{c}{Qwen3-8B} \\
\cmidrule(lr){2-3}\cmidrule(lr){4-5}\cmidrule(lr){6-7}\cmidrule(lr){8-9}
Task & 0-shot & Spec. & 0-shot & Spec. & 0-shot & Spec. & 0-shot & Spec. \\
\midrule
Banking77 & .4354 & .9302 & .5747 & .9276 & .6071 & .9312 & .6438 & .9321 \\
\quad (5-shot) & .5679 & & .6273 & & .6357 & & .6594 & \\
Spider & .5242 & .6963 & .6567 & .6992 & .6760 & .7321 & .6963 & .7582 \\
xLAM-FC & .5770 & .8205 & .6315 & .8225 & .7940 & .8480 & .7700 & .8455 \\
\bottomrule
\end{tabular}
\end{table}

\subsection{Retention across nine upgrade hops}
\label{sec:results:tax}

Table~\ref{tab:strategy_results} reports $R(\method{Freeze})$
(Eq.~\ref{eq:retention}) for every consecutive hop. On Banking77 the
frozen specialist retains 99--101\% of what retraining would deliver
on every hop. The retrained reference score is flat at 92.8--93.2\% across
four generations (Figure~\ref{fig:halflife}a) while zero-shot climbed
from 43.5\% to 64.4\%. On xLAM the pattern holds in two of three hops
(0.99, 1.03; neither the shortfall nor the excess is significant), with
one taxed hop (0.53) caused by Qwen2.5's unusually strong zero-shot
function calling. Spider is the systematic outlier:
freezing retains only 41--93\% (mean 0.64), because the retrain ceiling
itself rises with every generation (69.6$\to$75.8): new-base SQL
competence penetrates the fine-tune. One caution applies to these
single-hop values on the generative task: a second training seed for
the Spider/Qwen3 reference lands at .7331 rather than .7582, which moves
the final-hop $R(\method{Freeze})$ anywhere from 0.58 to 0.97 and ties
the freeze-vs-retrain comparison ($p{=}1.0$ against the second seed).
Single-hop retention on generative tasks is seed-limited at this test
size; the base-bound classification of Spider rests instead on the
lineage-level ceiling rise ($+5.9$pp over two hops, $p{<}10^{-7}$),
which no seed choice undoes.

\begin{table}[t]
\centering
\small
\caption{Retention by strategy. \method{Freeze} is measured on all
nine hops; \method{Copy} on the one shape-compatible hop
(2$\to$2.5); \method{Refresh-D} on the most recent hop (2.5$\to$3),
with cost = teacher relabeling + student training on one RTX 4090.
$R < 0$: worse than not transferring. $R > 1$: beats gold-label
retraining.}
\label{tab:strategy_results}
\begin{tabular}{lcccc|c|cc}
\toprule
 & \multicolumn{4}{c}{$R(\method{Freeze})$} &
\multicolumn{1}{c}{$R(\method{Copy})$} &
\multicolumn{2}{c}{\method{Refresh-D}} \\
\cmidrule(lr){2-5}\cmidrule(lr){6-6}\cmidrule(lr){7-8}
Task & 1.5$\to$2 & 2$\to$2.5 & 2.5$\to$3 & mean & 2$\to$2.5 & $R$ & cost (min) \\
\midrule
Banking77 & 1.007 & 0.989 & 0.997 & 0.998 & \textbf{--0.551} & 0.997 & 39 \\
Spider    & 0.932 & 0.414 & 0.578 & 0.641 & 0.139 & \textbf{1.047} & 255 \\
xLAM-FC   & 0.990 & 0.528 & 1.033 & 0.850 & 0.370 & 0.960 & 331 \\
\bottomrule
\end{tabular}
\end{table}

This pattern refutes our pre-registered H3 (D $>$ S $>$ A): measured
\method{Freeze} retention ranks D $\approx$ A $>$ S. What separates
the tasks is not surface form but \emph{competence provenance}: where
the skill is mostly \emph{data-bound}, as with a label taxonomy or a call
format, the fine-tune is the ceiling and base generations are
interchangeable. Where it is \emph{base-bound}, as in compositional SQL
over unseen schemas, base progress compounds with fine-tuning, and
freezing forfeits that progress. We propose competence provenance as the
organizing variable for transfer evaluations, in place of the
discriminative/generative split.

\subsection{Reference-score drift and task--base coupling}
\label{sec:results:ceilings}

Provenance makes a testable prediction: adding a second task per class
should sort by what the skill depends on, not by class. It does
(Table~\ref{tab:ceilings}). Across the three most recent 7--8B
generations, with three training seeds on every endpoint where the
claim depends on it, retrained reference scores rise on Spider
($+4.8$pp between seed means, .7025 $\to$ .7501; every Qwen3 seed
exceeds every Qwen2 seed) and, mildly, on xLAM-FC ($+2.3$pp single
seed; three seeds at each endpoint span $\le$0.8pp and confirm the
direction); they are flat on Banking77, glaive-FC, and FinQA. On
CLINC150, the seed-42 pair showed a significant decline ($-0.9$pp,
$p{=}0.002$), but three seeds per endpoint overturn it: the seed
ranges overlap (Qwen2 .9269--.9344, Qwen3 .9180--.9313) and the gap
between seed means ($-0.5$pp) is smaller than the within-generation
spread (1.3pp), so we report CLINC150 as flat within training noise,
not declining---an instance of exactly the single-seed fragility our
protocol is meant to catch. FinQA's flat
reference score despite being structured generation is consistent with its
skill being dominated by a dataset-specific program DSL (its
adoption floors are low and erratic across generations, .09--.22),
i.e., data-bound in our terms even though the surface form is
generative.

\begin{table}[t]
\centering
\small
\caption{Fine-tuned ceilings across the three most recent 7--8B
generations (specialist scores; reference for hops ending at each
generation). $\Delta$ = Qwen3 minus Qwen2; $p$ from the exact McNemar
test on the paired test set.}
\label{tab:ceilings}
\begin{tabular}{llccccl}
\toprule
Task & Class & Qwen2 & Qwen2.5 & Qwen3 & $\Delta$ (pp) & Evidence \\
\midrule
Spider$^{3s}$ & S & .7025 {\scriptsize[.695--.713]} & .7321 {\scriptsize[.730--.734]} & .7501 {\scriptsize[.740--.758]} & $+4.8$ & all seeds ordered \\
xLAM-FC$^{3s}$ & A & .8225 & .8473 {\scriptsize[.846--.848]} & .8460 {\scriptsize[.843--.850]} & $+2.3$ & $3{\times}10^{-4}$ \\
FinQA     & S & .5981 & .5990 & .6103 & $+1.2$ & $0.22$ \\
Banking77 & D & .9276 & .9312 & .9321 & $+0.5$ & $0.16$ \\
glaive-FC & A & .9640 & .9650 & .9645 & $+0.05$ & $1.0$ \\
CLINC150$^{3s}$ & D & .9313 {\scriptsize[.927--.934]} & .9322 & .9262 {\scriptsize[.918--.931]} & $-0.5$ & within seed noise \\
\bottomrule
\end{tabular}
\end{table}

A binary base-bound/data-bound label invites circularity, so we
summarize the coupling as a single elasticity per task:
$\beta_t = \Delta O_t / \Delta F_t$, the reference-score change per
point of zero-shot-floor change over the same hops (Qwen2$\to$3,
7--8B). The six tasks form a continuum rather than two classes:
Spider $\beta{=}1.49$, xLAM $0.17$, Banking77 $0.07$, glaive-FC
$0.00$, CLINC150 $-0.07$; FinQA's floor moves less than 1pp, so its
$\beta$ is denominator-unstable and we do not report one. We use
``high-'' and ``low-coupling'' as shorthand for the ends of this
scale, and \S\ref{sec:results:smallscale} shows the coupling itself
moves with model scale, so $\beta$ is a property of a
task--scale--release window, not of a task.

\subsection{Copying and the continuity criterion}
\label{sec:results:copy}

\begin{figure}[t]
\centering
\includegraphics[width=\linewidth]{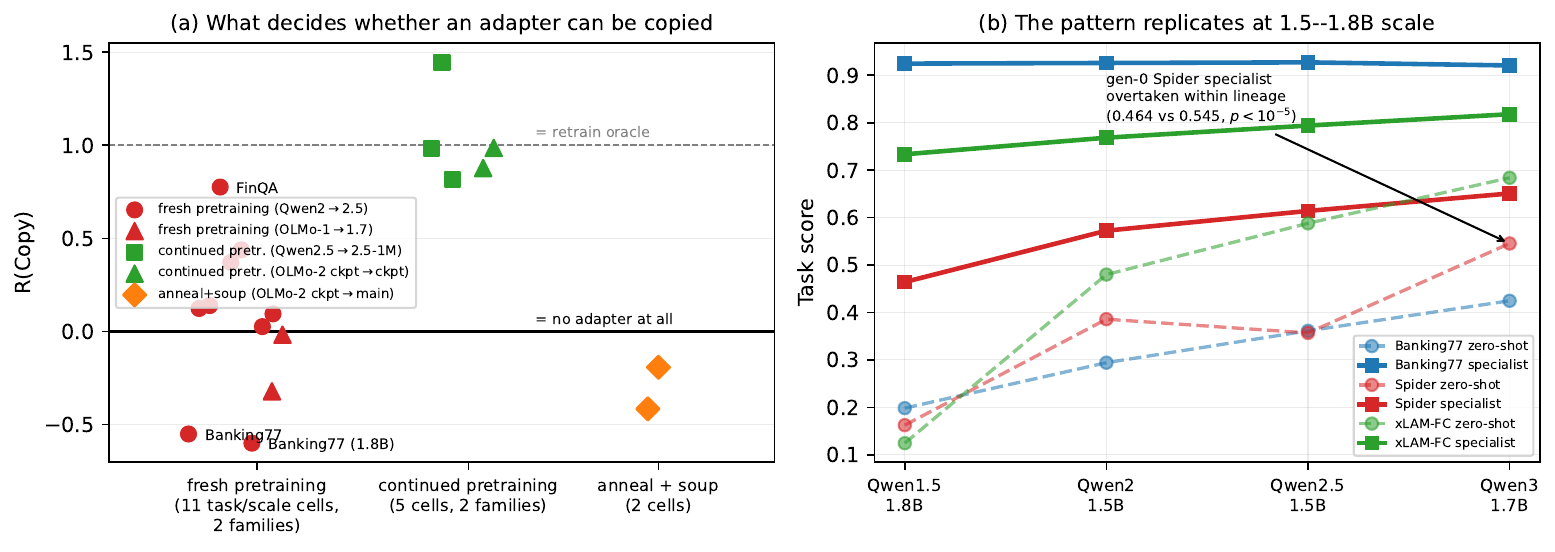}
\caption{\textbf{(a)}~Retention of naive adapter copying between
architecturally identical checkpoints, in two families. Between
independent pretraining runs (Qwen2$\to$2.5, nine task/scale cells;
OLMo-1$\to$1.7, two cells) copying lands at or below
the no-adapter floor; between checkpoints connected by continued
pretraining (Qwen2.5$\to$2.5-1M, three tasks; OLMo-2 stage-1
\texttt{237k}$\to$\texttt{248k}, two cells) it retains 0.82--1.45
(Spider's 1.45 reflects a small denominator: the 1M reference gains only
3.5pp over its floor); and once the continuation runs long
(\texttt{237k}$\to$main, ${\sim}$2.9T tokens plus anneal and soup) it
collapses below the floor---Table~\ref{tab:olmo}'s distance ablation
shows the 2.9T-token distance alone suffices, with the release
operations adding nothing further. \textbf{(b)}~The pattern
replicates at
1.5--1.8B: the Banking77 ceiling stays flat while zero-shot climbs
22.6pp, the Spider ceiling rises with every generation, and the
generation-0 Spider specialist is overtaken \emph{within} the lineage.}
\label{fig:copy}
\end{figure}

Two of three major-version hops in this lineage admit no weight-space
transfer at all (Table~\ref{tab:lineages}). The 2$\to$2.5 hop does:
the two checkpoints are identical in hidden width, depth, attention
geometry, FFN size, and vocabulary, so a LoRA adapter loads without any
mapping. Prior work describes this as the setting where naive copying
should work; our prospectively specified H2 predicted
$R \ge 0.8$ here.

In practice it is dominated by \method{Freeze} on every task we test,
and on most tasks it lands at or below the no-adapter floor
(Figure~\ref{fig:copy}a). On Banking77 the copied adapter scores
.4286 [.411,\,.446], a 50-point collapse from the source specialist's
.9276 and 17.9 points below the .6071 that the target base achieves
with no adapter at all, giving $R = -0.551$. The other five tasks land
between harm and marginal gain: CLINC150 $R{=}0.123$, Spider
$R{=}0.139$, xLAM $R{=}0.370$, glaive-FC $R{=}0.69$ (on the
re-split described in \S\ref{sec:tasks}; copy .946 vs.\ floor .903
vs.\ reference .965, above the floor, $p{=}2{\times}10^{-15}$, yet
below both the reference and \method{Freeze},
$p{\le}6{\times}10^{-5}$), and FinQA
$R{=}0.776$ (inflated by that task's
anomalously low adoption floor of .0933). At 1.5--1.8B scale the same
hop gives $R{=}-0.600$ (the Banking77 adapter collapses to 2.2\%
absolute), $0.026$, and $0.095$. Across nine task/scale cells, seven
lie at or below $R{=}0.37$; the two exceptions (glaive-FC 0.69, FinQA
0.78) owe their size to format-heavy tasks with shallow adapters and
still trail their \method{Freeze} retentions. In
every cell copying is worse than doing nothing to the old model
(\method{Freeze} spans 0.41--1.04 on the same hop).

The continuation hop reverses the verdict. Qwen2.5-7B-Instruct-1M is a
continued-pretraining descendant of the same weights, and copying the
same three adapters onto it retains $R{=}0.985$ (Banking77), $1.445$
(Spider; the raw scores are copy .7292 vs.\ floor .6789 vs.\ reference
.7137, so the ratio's denominator is small), and $0.815$ (xLAM). On
all three tasks no difference from full retraining is detected
($p{=}0.15$, $0.16$, $0.07$) while the copy sits far above the
adoption floor ($p{\le}4{\times}10^{-4}$); but absence of detection
is not parity, so we also test non-inferiority against pre-set margins
(1pp classification, 2pp structured) using paired bootstrap CIs and
three reference seeds. Banking77 is non-inferior at 2pp but borderline
at 1pp (copy$-$reference CI $[-1.01, +0.10]$pp; three-seed reference
band .9308--.9325). Spider's copy (.7292) lies \emph{inside} the
three-seed reference range (.7137/.7340/.7408), so the comparison is
bounded by training noise rather than by item noise. xLAM fails the
1pp test outright (CI $[-2.00, 0.00]$). In short:
copying onto a short continuation costs at most about a point;
copying across a fresh run costs fifty. Architectural compatibility is the wrong
test; a documented, \emph{short} continued-training path is the right
one, which also explains
why within-family porting studies
\citep{huang2024chatvector,lin2025finetuningtransfer} report success:
their hops are continuations. Incidentally, the 1M release itself buys
specialists nothing: frozen 2.5 specialists match or exceed 1M-retrained
references on all three tasks (e.g., Spider .7321 vs.\ .7137,
$p{=}0.10$).

\begin{table}[t]
\centering
\small
\caption{OLMo panel: copying the same intent adapters along a fully
documented training trajectory. The top pair is a verified fresh
pretraining run; the bottom four rows per task move the target along
OLMo-2's own trajectory---a 46B-token continuation, a 2.9T-token
continuation (end of stage 1, before any annealing), the stage-2
anneal (before souping), and the released soup. Copy vs.\ floor and
vs.\ reference tested with exact McNemar ($\downarrow$: \method{Copy}
significantly below the floor; $\approx$: indistinguishable).
CLINC150 uses the lenient metric (strict released per cell).}
\label{tab:olmo}
\begin{tabular}{llrrrrl}
\toprule
Target edge & Task & \method{Copy} & Floor & Ref & $R$ & $p$ vs floor \\
\midrule
OLMo-1 $\to$ 1.7 \emph{(fresh run)} & B77 & .0266 & .2464 & .9292 & $-0.32$ & $<\!10^{-169}$ ($\downarrow$) \\
 & CLINC & .2364 & .2482 & .9151 & $-0.02$ & $0.063$ ($\approx$) \\
\midrule
237k $\to$ 248k \emph{(+46B cont.)} & B77 & .8179 & .0097 & .9305 & $+0.88$ & $<\!10^{-300}$ \\
 & CLINC & .8951 & .0120 & .9078 & $+0.99$ & $<\!10^{-300}$ \\
237k $\to$ 928k \emph{(+2.9T cont.)} & B77 & .1279 & .1432 & .9289 & $-0.02$ & $0.074$ ($\approx$) \\
 & CLINC & .1433 & .0315 & .9204 & $+0.13$ & $<\!10^{-138}$ \\
237k $\to$ ing.\ 3 \emph{(+anneal)} & B77 & .1276 & .3055 & .9315 & $-0.28$ & $<\!10^{-78}$ ($\downarrow$) \\
 & CLINC & .1296 & .2822 & .9122 & $-0.24$ & $<\!10^{-111}$ ($\downarrow$) \\
237k $\to$ main \emph{(+soup)} & B77 & .1140 & .3523 & .9256 & $-0.42$ & $<\!10^{-118}$ ($\downarrow$) \\
 & CLINC & .1313 & .2575 & .9125 & $-0.19$ & $<\!10^{-78}$ ($\downarrow$) \\
\bottomrule
\end{tabular}
\end{table}

\textbf{Ground-truth genealogy: portability decays with distance.} The
Qwen evidence rests on inferred lineage; OLMo makes the training
relationship an experimental variable (\S\ref{sec:lineages}), with
predictions fixed before each wave (\S\ref{sec:rq}).
Table~\ref{tab:olmo} shows the ten cells. On the verified fresh pair
(OLMo-1 $\to$ OLMo-1.7, identical shapes, independent runs), copying
collapses exactly as on Qwen2$\to$2.5: to 22.0 points \emph{below} the
target's no-adapter floor on Banking77 ($R{=}-0.32$) and to statistical
indistinguishability from the floor on CLINC150 ($R{=}-0.02$,
$p{=}0.063$), against references above .91.

\begin{figure}[t]
\centering
\includegraphics[width=0.72\linewidth]{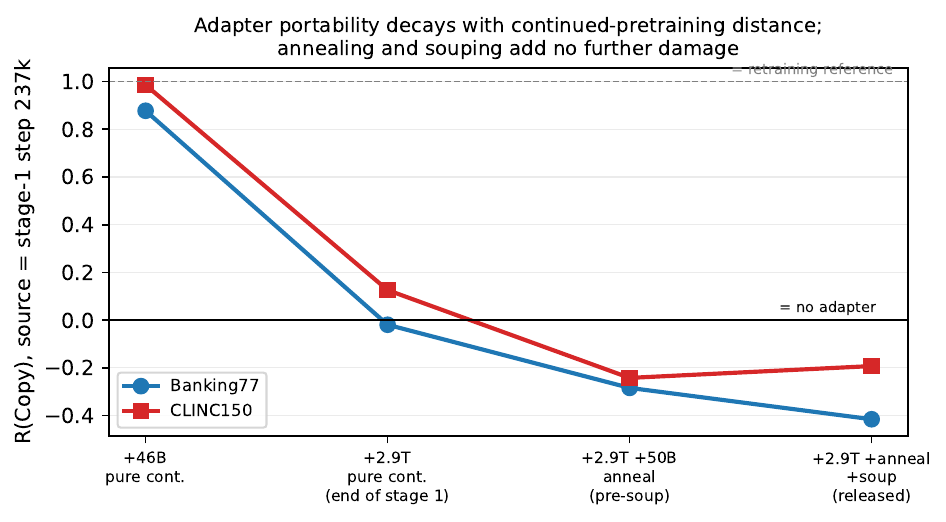}
\caption{Copying one adapter (trained at OLMo-2 stage-1 step 237k)
onto four later checkpoints of the same documented run. Retention
falls to the no-adapter floor within the pure-continuation segment;
the anneal and soup edges change the copied score by at most 1.4pp
(the lower $R$ reflects their higher floors). Generated from the
released records.}
\label{fig:spectrum}
\end{figure}

The four-edge spectrum then holds the source fixed (the stage-1
checkpoint at 237k steps, ${\sim}1$T tokens in) and moves the target
along the run's own documented trajectory. At a 46B-token continuation
(${\sim}1\%$ of the run) the copied adapter retains $R{=}0.88$ and
$0.99$---though exact parity with the reference is already rejected on
Banking77 ($-11.3$pp, $p{<}10^{-79}$), so even short continuations
exact a measurable tax. At the end of stage 1---2.9T further tokens of
\emph{pure} continuation, no annealing, no merging---the same adapter
is statistically at the floor on Banking77 ($R{=}-0.02$) and retains a
small residual signal on CLINC150 ($R{=}+0.13$). Crossing the stage-2
anneal and the final model-soup merge changes the copied score barely
at all (Copy stays between .11 and .14 on both tasks across all three
far edges); the more negative $R$ values there reflect the rising
zero-shot floors of later checkpoints, not additional damage from the
release operations. Our wave-4 text, written when only the endpoints
existed, attributed the collapse to the anneal-plus-soup boundary; the
distance-matched ablation overturns that reading, and we report the
correction (\S\ref{sec:rq}). \emph{Continuation distance is the
operative variable}: an adapter's portability comes with a drift
budget measured in continued-pretraining tokens---high retention at
1\% of the run, zero at 74\%---and annealing or souping at the end of
that distance neither rescues nor further destroys it. This is also
consistent with the Qwen continuation result: the 1M long-context
release is a short continuation of Qwen2.5, and copying onto it
retains 0.82--1.45. Figure~\ref{fig:spectrum} plots the spectrum.

The failure ordering on the fresh hop---A $>$ S $>$ D---inverts the
\method{Freeze} ordering: the adapters that survive copying best are
the ones that matter least, while the classification adapters
delivering the largest frozen advantage are the ones copying
destroys.

Four checks rule out the obvious artifacts. First, the failure is
not a decoding or parsing problem: of the copied model's 3{,}080
outputs, 48.1\% are near-miss label variants absent from the taxonomy
(\texttt{card\_not\_arrived} where the gold label is
\texttt{card\_arrival}), with no empty or degenerate generations.
Second, it is not a chat-template mismatch: a full 2$\times$2 over
\{source, target\} template $\times$ \{source, target\} base with
the same adapter leaves every score unchanged to four decimal places
(.4286 on the target base under either template; .9276 on the source
base under either template)---the base--adapter pairing alone
determines the outcome. Third, it is not a 4-bit artifact: re-running
the copy in bf16 without quantization on the first 512 test items
gives 52.5\% against 52.0\% quantized (per-example records
released), a 0.6pp difference against a 50-point collapse. Fourth, it
is not purely an output-format regression: with decoding constrained
to the 151 valid CLINC labels, the copied adapter recovers from .7505
to .8669 but remains 7.4pp below the constrained retrained reference
(.9413), so a real mapping loss persists once formatting is removed
from the equation. Constrained decoding also shrinks the CLINC
generational decline (\S\ref{sec:results:tax}) from $-0.89$pp to
$-0.65$pp: roughly a third of that headline number is formatting, not
competence.

\textbf{Rank dependence.} All results above use the fixed $r{=}16$
recipe. Re-running the fresh Qwen2$\to$2.5 hop at $r{=}64$ (adapter,
reference, and copy all at the higher rank) weakens the collapse
substantially: the Banking77 copy scores .7205 against a .9266
reference ($R{=}+0.29$, vs.\ $-0.55$ at $r{=}16$), and the Spider
copy reaches .7302 against a .7495 reference ($R{=}0.74$, vs.\ 0.14).
Higher rank also lifts the Spider reference itself (+4.7pp over
$r{=}16$). The ordering that drives the decision---copy below
\method{Freeze} and below retraining---holds at both ranks, but the
collapse \emph{magnitude} is a function of adapter rank, so the
$r{=}16$ numbers are one point on that axis, not the transfer profile
of fine-tuning in general (full fine-tuning remains unmeasured;
single seed, one hop, two tasks).

A plausible mechanism is that a 77-way label taxonomy is encoded as
sharp, low-entropy surgery on the output distribution, making it
maximally sensitive to drift in the underlying residual stream, whereas
format-and-style adaptations such as JSON call syntax or SQL conventions
survive drift because they are shallower. We have not tested this
directly, and layer-wise ablation of what each adapter changes is
future work. The actionable result is the criterion: before porting an
adapter, ask (i)~whether the target is reachable from the source base
by a documented continued-training trajectory at all---fresh
initializations collapse immediately---and (ii)~\emph{how far} along
that trajectory it sits, because retention decays from near-parity at
a 46B-token continuation to nothing by 2.9T tokens. The answer is in
the release genealogy, not in \texttt{config.json}, and a copied
adapter should in any case pass a task-level regression gate before
serving.

\subsection{Measured durability horizons}
\label{sec:results:halflife}

We track the February-2024 Spider specialist against each later base's
cheap adoption, in real release time. Its advantage starts at
$+17.2$pp and is $+4.0$pp four months later at $g{=}1$ (McNemar
$p{=}0.005$)---77\% of the initial advantage is gone within one
release interval, so $H_{50} \in (0, 4]$ months. The crossover takes
far longer: $+2.0$pp at seven months ($p{=}0.19$, unresolvable) and
exactly $0.0$pp at fourteen ($g{=}3$: the two systems disagree on 234
of 1{,}034 items and split them 117 each, $p{=}1.0$), giving
$H_0 = 14$ months, censored to the interval $(7, 14]$. The two
horizons differ by an order of magnitude, which a single ``half-life''
number would conflate; and reading a point estimate off the $g{=}3$
tie would overstate what a 1{,}034-item test set can resolve.

The Banking77 specialist from the same date behaves differently: it
retains $+28.6$pp at fourteen months ($p<10^{-200}$: 923 items it
answers correctly and Qwen3 zero-shot does not, against 41 the other
way), with mean decay 2.9pp per generation, so $H_{50}$ and $H_0$ both
lie beyond the observed series. The xLAM specialist crosses its
half-value point between four and seven months
($H_{50} \in (4, 7]$) yet still retains $+5.1$pp at fourteen
($p<10^{-19}$), non-monotonically (\S\ref{sec:results:nonmono}).

H1 (prospective) predicted crossover within one generation for
reasoning-heavy structured tasks and two or more for discriminative
ones. The discriminative clause holds; the structured clause is wrong
about the crossover ($H_0$ took the whole series) but right in spirit
about value decay ($H_{50}$ under one release interval). Even on the
most base-coupled task in our suite, specialization out-earned cheap
adoption for over a year---while losing most of its \emph{margin}
within four months.

\textbf{A second family.} Repeating the Spider durability measurement
on the OLMo release series (OLMo-1 $\to$ 1.7 $\to$ 2; February to
November 2024; base checkpoints, plain format, floors and references
under the identical format) gives the opposite trajectory: the frozen
OLMo-1 specialist (.6673) holds a $+48.4$pp advantage at release,
$+39.4$pp two months later, and $+58.1$pp at nine months---no
half-value crossing at all---because this series' zero-shot SQL floors
stay between .09 and .27 and its retrained references are flat
(.667/.685/.668). The same task whose margin halved within four months
on Qwen loses nothing in nine months on OLMo. This refutes our W1
prediction and sharpens the durability finding: erosion is driven by
the release series' floor growth, not by the task; $H_{50}$ and $H_0$
are properties of the task--series pair. (Within-family comparison
only: OLMo floors are base-model floors and are not comparable to
Qwen's instruct floors in absolute terms.)

\subsection{Non-monotonic generational uplift}
\label{sec:results:nonmono}

Zero-shot uplift per generation is task-dependent in \emph{sign}:
xLAM-FC drops 2.4pp from Qwen2.5 to Qwen3 (.7940 $\to$ .7700), so the
2024 specialist's advantage \emph{rebounds} from +2.6pp at $g{=}2$ to
+5.1pp at $g{=}3$ (Figure~\ref{fig:halflife}b). Averages like the
densing law \citep{xiao2024densing} govern the trend, not any single
task: ``wait for the next base'' is not directionally safe without
task-level measurement, which is precisely what the benchmark's
released harness automates.

\subsection{Replication at 1.5--1.8B scale}
\label{sec:results:smallscale}

Running the whole lineage again with 1.5--1.8B checkpoints reproduces
the structure (Figure~\ref{fig:copy}b). The Banking77 ceiling is
statistically flat across four generations (.9247 $\to$ .9208,
$p{=}0.32$) while zero-shot climbs from .1984 to .4247; the gen-0
specialist still leads the newest base by 50 points. The Spider ceiling
rises every generation (.4642 $\to$ .6509, $p{<}10^{-26}$), and here
the half-life resolves inside the lineage: the gen-0 specialist's
advantage is positive through $g{=}2$ but $-8.1$pp at $g{=}3$ (.4642
vs.\ .5455, $p{=}2{\times}10^{-6}$). At 7--8B we could only bracket
the crossing; at 1.5--1.8B it is directly observed, on the same task,
in the same direction.

The replication also sharpens the taxonomy in one respect: xLAM-FC,
whose 7--8B ceiling is nearly flat ($+2.3$pp), gains $+8.5$pp across
the small-scale lineage ($p{=}3{\times}10^{-27}$). Competence
provenance is therefore not a fixed attribute of a task but a relation
between the task and the model's capability: while a model class is
still far from saturating a task, base progress penetrates fine-tuning
(base-bound behavior); once the task is saturated at that scale, the
fine-tune becomes the ceiling (data-bound behavior). This predicts
that today's base-bound tasks will migrate to data-bound as bases
improve, which is itself a measurable, falsifiable claim the released
harness can track.

\subsection{Label-free distillation refresh}
\label{sec:results:refresh}

\method{Refresh-D} is \emph{annotation-free but input-retaining}:
the old specialist relabels the retained task \emph{inputs}, and a
fresh LoRA is trained on the new base (it is not ``data-free''---the
inputs must have been kept). Teacher--gold agreement is 97.5\%
(Banking77 labels), 88.3\% (Spider, exact-string, which
underestimates semantic agreement), and 92.8\% (xLAM). Refreshed
students score .9312 / .7611 / .8425, i.e., retention 0.997 / 1.047 /
0.960 against gold-label retraining, with no detected difference
(McNemar $p{=}0.81$, $0.82$, $0.58$). Against the pre-set margins,
Banking77 is formally non-inferior at 1pp (paired CI
$[-0.62, +0.42]$pp; three student seeds span .9302--.9312); Spider and
xLAM are non-inferior at 2pp but not at 1pp (CIs $[-1.45, +1.93]$ and
$[-1.20, +0.60]$). Spider's student seeds span .7302--.7689 (3.9pp),
so its single-run point estimate sitting above the reference (.7611
vs.\ .7582) is inside training noise; a teacher that is itself only
73\% accurate producing training data of reference quality would echo
weak-to-strong generalization \citep{burns2023weak}, but our three
seeds cannot separate that from chance.

A fourth cell added at review extends the pattern to a second task
class: a CLINC150 student distilled from the Qwen2.5 specialist scores
.9267 on Qwen3, inside the three-seed band of its gold-label
references (.9180--.9313). Where copying was applicable and achieved
positive retention, refresh
delivered 2.6$\times$ (xLAM) to 7.5$\times$ (Spider) more of the
attainable gain; on Banking77 the comparison is degenerate because
copying was net harmful. The resource picture, however, is
asymmetric: counting teacher inference, refresh costs 39.1 GPU-minutes
on Banking77 (vs.\ 45.9 for gold retraining) but 255.0 on Spider
(vs.\ 205.8) and 331.2 on xLAM (vs.\ 215.8). \method{Refresh-D}
saves \emph{annotation}, not compute; and if the original gold labels
were retained alongside the inputs, plain retraining dominates it.
Zero gold labels, zero API calls.

\subsection{Recovery cost under a label budget}
\label{sec:results:ladder}

On the taxed task (Spider, Qwen3), gold-label retraining with a budget
ladder yields .6847 (256 labels---nominally below the .6963 zero-shot
floor, though the gap is not significant, $p{=}0.46$: 256 labels buy
nothing), .7205 (1{,}024; 39\% of attainable gain), .7427 (4{,}096;
75\%), .7582 (full 6{,}362). The
95\%-of-reference-gain budget $C_{95}$ exceeds 4{,}096 labels, whereas label-free
\method{Refresh-D} reaches 105\% using retained inputs only. Where the upgrade tax is real, the cheapest way to pay it is to
recycle the old specialist's behavior, not to re-annotate.

\subsection{Backward compatibility: what averages hide}
\label{sec:results:flips}

An upgrade that leaves mean accuracy unchanged can still regress
previously verified behavior. From the released per-example records we
compute, for every strategy pair on the identical test set, the
\emph{negative flip rate} (items the old system answered correctly and
the new one does not) alongside the positive flip rate
(Table~\ref{tab:flips}). Three results matter for deployment.
First, even quality-neutral retraining churns: replacing the Qwen2.5
specialist with a retrained Qwen3 reference flips 1.2\% (Banking77) to
4.7\% (Spider) of items negatively even when the net change is zero or
positive. Second, the CLINC150 decline of $-0.89$pp decomposes into
2.55\% negative against 1.65\% positive flips---140 previously-correct
items break, invisible in the mean. Third, naive
copying is catastrophic on this axis: the copied Banking77 adapter
negatively flips \emph{half the test set} (50.4\%) relative to the
source specialist. \method{Refresh-D} shows the same flip profile as
gold retraining (1.1--3.8\% negative), so annotation-free refresh does
not add churn beyond retraining itself.

\begin{table}[t]
\centering
\small
\caption{Negative/positive flip rates between strategy pairs on the
fixed test sets (per-example records; NF = old correct, new wrong).}
\label{tab:flips}
\begin{tabular}{lrrr}
\toprule
Transition & NF\% & PF\% & Net (pp) \\
\midrule
Banking77: 2.5-specialist $\to$ 3-reference & 1.23 & 1.33 & $+0.10$ \\
Spider: 2.5-specialist $\to$ 3-reference & 4.74 & 7.35 & $+2.61$ \\
xLAM: 2.5-specialist $\to$ 3-reference & 3.90 & 3.65 & $-0.25$ \\
CLINC150: 2-specialist $\to$ 3-specialist & 2.55 & 1.65 & $-0.89$ \\
FinQA: 2-specialist $\to$ 3-specialist & 4.27 & 5.49 & $+1.22$ \\
glaive-FC: 2-specialist $\to$ 3-specialist & 0.50 & 0.40 & $-0.10$ \\
Banking77: source specialist $\to$ \method{Copy} & 50.39 & 0.49 & $-49.90$ \\
Banking77: reference $\to$ \method{Refresh-D} & 1.17 & 1.07 & $-0.10$ \\
Spider: reference $\to$ \method{Refresh-D} & 3.77 & 4.06 & $+0.29$ \\
xLAM: reference $\to$ \method{Refresh-D} & 2.20 & 1.90 & $-0.30$ \\
\bottomrule
\end{tabular}
\end{table}

\subsection{Does the decision policy beat fixed strategies?}
\label{sec:results:policy}

The framework's claim is that genealogy, coupling, and flips suffice to
\emph{choose} well. We test this by offline replay over the 33
task$\times$scale$\times$hop episodes in which at least
\method{Freeze}, \method{Adopt}, and the retraining reference are all
measured (Copy on 18 of them, \method{Refresh-D} on 3). The
pre-specified policy is: (i)~if the reference beats \method{Freeze} by
at most $\varepsilon$ (1pp classification, 2pp structured), freeze;
(ii)~else if a copy is defined, the target is a documented short
continuation, and the copy passes a regression gate at $\varepsilon$,
copy; (iii)~else if retained inputs exist and the refreshed student
passes the gate, refresh; (iv)~else retrain. Each policy is charged
the training cost of the action it picks and its regret against the
best \emph{measured} action of that episode
(Table~\ref{tab:policy}).

\begin{table}[t]
\centering
\small
\caption{Offline policy replay over 33 measured upgrade episodes.
Regret is against the best measured action per episode; a regression
episode is one where the chosen action falls more than $\varepsilon$
below \method{Freeze}.}
\label{tab:policy}
\begin{tabular}{lrrrrr}
\toprule
Policy & Mean regret & Max regret & GPU-h & Gold-label eps. & Regressions \\
\midrule
Always-\method{Freeze} & 1.80pp & 11.5pp & 0 & 0 & 0 \\
Always-\method{Adopt} & 32.35pp & 91.6pp & 0 & 0 & 33 \\
Always-retrain & 0.21pp & 2.0pp & 47.7 & 33 & 1 \\
Copy-if-shape & 17.11pp & 90.8pp & 25.8 & 15 & 14 \\
Proposed & \textbf{0.37pp} & \textbf{6.6pp} & \textbf{15.6} & \textbf{9} & \textbf{0} \\
\bottomrule
\end{tabular}
\end{table}

To avoid scoring the policy on the same measurements its gates
consult, each test set is split in half by a hash of the item id: gate
decisions read only the gate half, and all regret and regression
figures are computed only on the held-out half. Under this split the
proposed policy tracks always-retraining (0.37 vs.\ 0.21pp mean
regret) at 33\% of its training compute and 27\% of its gold-label
episodes, with zero regression episodes; its worst episode (6.6pp)
comes from a gate miscall on a half-sized test set---gate sampling
error made visible. (Letting the gates read the full test set, as an
optimistic upper bound, gives 0.13pp.) The
instructive failure is \emph{copy-if-shape}---the rule practitioners
actually use: 17pp mean regret and 14 regression episodes, because
shape compatibility fires exactly where copying is most dangerous.
One caveat stands: regret is relative to \emph{measured} actions
only---an unmeasured method could dominate.

\subsection{Predicting portability before paying for it}
\label{sec:results:probe}

The decision framework still requires one measured copy per candidate
edge. RQ5 asks whether a probe that costs a single forward pass per
prompt can stand in for that measurement. For each of the eight base
pairs with a measured $R(\method{Copy})$, we run 256 fixed task
prompts through both checkpoints (no training, no generation) and
compute three signals: mean linear CKA between hidden states at nine
evenly spaced layers, next-token top-1 agreement, and next-token
Jensen--Shannon divergence. CKA rank-correlates with the pair's mean
$R(\method{Copy})$ at Spearman $\rho{=}0.74$ (Pearson 0.54); top-1
agreement and JSD are weaker (0.38, $-0.52$). The failure modes are
informative: CKA tracks the OLMo distance spectrum monotonically
(0.80 at 46B tokens down to 0.53 at the far edges) but assigns nearly
identical scores to the Qwen fresh pair and the Qwen continuation pair
(.876 vs.\ .878), whose retentions differ by 0.8; top-1 agreement
separates the Qwen pairs no better and is distorted by tokenizer
differences across families. With eight pairs this is an exploratory
result, not a decision rule: a probe carries real signal about
weight-space drift, but none of the three is a substitute for one
cheap measured copy plus a regression gate. We release the probe
harness so later waves can grow $n$.

\subsection{Reliability, variance, and threats}
\label{sec:results:threats}

\textbf{Format reliability.} Fine-tuning buys parseability as well as
accuracy: unparseable xLAM outputs drop from 17--28/2000 (zero-shot)
to 5--6/2000 (specialists); copied adapters land in between
(15/2000). Banking77 parsing failures are zero in all conditions.
\textbf{Variance.} Three seeds on the Banking77/Qwen2.5 cell give
.9312/.9289/.9302 ($\sigma \approx 0.12$pp); three seeds on the
Spider/Qwen3 reference give .7582/.7398/.7524, a 1.8pp spread that
dominates single-hop retention arithmetic on that task
(\S\ref{sec:results:tax}). Bootstrap CIs accompany every cell in the
release. After wave 6, every generational reference-score claim the
paper leans on carries three seeds at the endpoints in question
(Spider, CLINC150, xLAM); comparisons still resting on single seeds
are those whose differences we already report as null.
\textbf{Threats.} One lineage (family culture and generational effects
are confounded); one dataset per task class; Spider uses
single-database execution accuracy; xLAM-FC is an in-distribution
held-out split that emphasizes format compliance; the Qwen3 hop
changes parameter count (7.6B$\to$8.2B) and data vintage together.
UpgradeBench prices the upgrade decision as practitioners face it,
bundled, rather than isolating the effect of any single factor.
\method{Copy} is measured on two fresh-pretraining hops (two
families, two scales), four documented continuation edges spanning
46B--2.9T tokens, one anneal edge, and one soup edge; the wave-5
distance ablation separates distance from the release operations at
these endpoints, but fitting a full decay curve needs more
intermediate points. The OLMo
cells use base checkpoints with a fixed plain prompt format on the two
intent tasks, so cross-family comparisons of absolute scores are not
meaningful; all wave-4 conclusions are within-family, against floors
and references measured under the identical format.
The memorization probe (\S\ref{sec:tasks},
Appendix~\ref{app:contamination}) addresses contamination; residual
contamination would inflate baselines and thus \emph{understate}
expert advantages.

% =====================================================================
\section{The Small-Evidence Regime}
\label{sec:smalln}

Every analysis above runs at $n{=}1{,}034$--$5{,}500$ paired test items.
An organization applying the decision policy to its own specialist
typically holds $n{=}100$--$600$ labeled items, and on a saturated task
the two systems disagree on only a handful of them. This section closes
that external-validity gap: we characterize the policy's operating
behavior at enterprise evidence sizes by replaying it over the
benchmark's own per-example records, and we revise the opportunity gate
so that its verdicts remain calibrated when the evidence cannot carry a
point estimate.

\textbf{A three-zone opportunity gate.} At small $n$ a point estimate
passed through a fixed margin conflates three different epistemic
states. We therefore replace it with an equivalence-testing (TOST-style)
verdict on the paired-difference confidence interval, computed from the
discordant counts $(n_{01}, n_{10})$ with a continuity correction, and a
rule-of-three bound $|\delta| \le 3.69/n$ when the two systems agree on
every item. The CI lying entirely above $\varepsilon$ \emph{establishes}
the gain and opens the waterfall of \S\ref{sec:results:policy}; entirely below
$\varepsilon$ establishes \emph{equivalence} (freeze as a verdict, with
the exclusion bound stated); an interval straddling $\varepsilon$ is
\emph{unresolved}, and the policy says so instead of defaulting to
freeze. Unresolved verdicts carry a posterior over the true gain, with
the benchmark's own 193-cell corpus as an empirical-Bayes prior per
lineage kind (fresh pretraining: $\mu{=}{+}0.82$pp, $\sigma{=}2.4$pp
over 16 pair-task samples; continuation: $\mu{=}{+}0.51$pp,
$\sigma{=}0.66$pp; seed-to-seed training variance from the multi-seed
waves is folded into the observation noise), and resolve to one of two
actions: \textsc{collect} (label the disagreement set) when the
posterior leaves the gain live, or \textsc{wait} (hold until the next
release) when $P(\text{gain} > \varepsilon_{\text{dec}}) < 0.10$.

\textbf{Disagreement-first labeling.} The exact McNemar test conditions
on the discordant pairs, so the decision-relevant annotation complexity
is $O(\#\text{disagreements})$, not $O(n)$: an organization can run both
systems over unlabeled traffic (free of annotation), collect the items
where their predictions differ, and label only those.

\textbf{Protocol.} We take the 21 (task, upgrade-pair) cells of the
main study for which both endpoints have full per-example records
(15 ground-truth null or negative at full $n$, i.e.\\
$\delta_{\text{full}} \le \varepsilon$; 6 ground-truth upgrades),
subsample enterprise-sized gate sets ($n \in \{100, \dots, 1000\}$,
300 replicates per cell per size, fixed seed), and score each verdict
against the full-$n$ ground truth. The corpus prior is computed
leave-one-pair-out, so no cell is judged under a prior that contains
itself. \textsc{collect} is played out, not assumed: up to 50 real
disagreement records are drawn and labeled, and the waterfall opens only
if the exact sign test settles positive. The comparison rule
(``old'') is the point-estimate gate of \S\ref{sec:results:policy} applied
directly at small $n$.

\begin{table}[t]
\centering
\small
\begin{tabular}{rccc}
\toprule
$n$ & \multicolumn{2}{c}{false-open (15 null cells)} &
false-equivalence \\
 & point-estimate gate & three-zone + \textsc{collect} & (6 upgrade
cells) \\
\midrule
100  & 17.9\% & 1.9\% & 0.5\% \\
200  & 15.2\% & 2.1\% & 0.2\% \\
300  & 13.0\% & 2.0\% & 0.2\% \\
600  & 8.2\%  & 1.8\% & 0.1\% \\
1000 & 1.8\%  & 1.5\% & 0.0\% \\
\bottomrule
\end{tabular}
\caption{Small-evidence operating curve, replayed on real records. At
enterprise gate sizes the point-estimate gate opens the upgrade
waterfall on one in six ground-truth-null episodes; the three-zone gate
holds the false-open rate near its nominal level at every $n$, and its
``equivalence'' verdicts are almost never wrong. The cost of that
conservatism is bounded and we state it: mean serving-accuracy regret
0.58pp vs 0.48pp at $n{=}100$ (the point-estimate rule's spurious opens
occasionally land on real upgrades; its migration and regression costs
are outside this metric).}
\label{tab:smalln-curve}
\end{table}

\begin{table}[t]
\centering
\small
\begin{tabular}{rcc}
\toprule
labeling budget $b$ & i.i.d.\ sampling & disagreement-first \\
\midrule
25  & 0\%  & 37\% \\
50  & 4\%  & 48\% \\
100 & 13\% & 67\% \\
\bottomrule
\end{tabular}
\caption{Probability that the upgrade direction settles (exact sign
test, correct sign) under a fixed labeling budget, averaged over the
cells with $|\delta_{\text{full}}| \ge 1$pp. Labeling $b$ random items
wastes almost the entire budget on items both systems already get
right; labeling $b$ disagreements concentrates it on the only items
that carry decision information.}
\label{tab:smalln-labels}
\end{table}

Table~\ref{tab:smalln-curve} shows the operating curve and
Table~\ref{tab:smalln-labels} the annotation-efficiency comparison. Three
regime-level facts follow. First, at $n \le 300$ the point-estimate gate
is not a usable instrument: its false-open rate (13--18\%) means roughly
one in six null episodes triggers a retrain-and-migrate cycle. Second,
honesty is cheap: refusing to over-claim costs about a tenth of a
percentage point of mean serving accuracy while cutting false opens by
an order of magnitude, and the equivalence verdicts that replace
``freeze by default'' are correct essentially always. Third, when the
verdict is unresolved, the efficient next step is never ``collect more
i.i.d.\ labels'': at these disagreement rates (1--15\%), a budget of 25
disagreement labels outperforms 100 i.i.d.\ labels by a factor of three
on direction resolution. The replay code and per-cell results ship with
our released tooling.

% =====================================================================
\section{Discussion}
\label{sec:discussion}

\textbf{Implications.} Our measurements are inconsistent with all three
policies an organization might adopt by default. ``Always retrain
on the newest base'' wastes 17--422 GPU-minutes per specialist per
generation for no measurable accuracy on data-bound tasks; on xLAM the
newest reference (.8455) does not even nominally exceed the
one-generation-old specialist (.8480). ``Never revisit fine-tunes'' forfeits 42--59\% of
attainable gain per hop on base-bound tasks. And ``just move the adapter over'' depends entirely on what the new
release is: harmful between independently pretrained checkpoints even
at identical architecture, near-free onto a short continuation of the
same weights, and back to harmful once the continuation runs long
(floor-level after 2.9T tokens, with annealing and souping adding
nothing further). Release genealogy plus distance, not
\texttt{config.json}, is what an upgrade policy should
read---and since released checkpoints typically sit trillions of
tokens past any point where an old adapter was trained, the copyable
regime is narrower than ``same family'' suggests. The
rational policy is per-task and per-release: check genealogy and
distance to
decide whether copying is even on the table, measure the task--base
coupling to decide between freezing and refreshing, and note that
the coupling itself drifts with scale and time
(\S\ref{sec:results:smallscale}). The refresh need not cost gold
labels, though it does not always save compute
(\S\ref{sec:results:refresh}). For the specialist-fleet debate
\citep{belcak2025slm}, this reframes staleness from an argument
against fine-tuning into a priceable line item: the durable assets are
the task data, the per-example regression records, and the refresh
pipeline, not the frozen weights. Freezing is not free either---it
carries security-patch, dependency, and compliance costs that our
protocol does not price.

\textbf{Future waves.} The released harness makes four extensions
mechanical: (i)~a further instruct-substrate lineage with a tokenizer
break (Llama~2$\to$3$\to$3.1; those checkpoints are license-gated,
which kept them out of this fully-automated study), and a finer-grained
sweep of intermediate OLMo checkpoints to fit the retention-vs-distance
decay curve whose endpoints we measured; (ii)~a versioned
synthetic private-knowledge track whose corpus is provably absent from
any pretraining data; (iii)~learned mappings
\citep{li2025lorasuite,xia2025crosslora} on the shape-incompatible
hops, where \method{Copy} is undefined and \method{Refresh-D} is
currently the only measured option; (iv)~representational probes (CKA,
logit agreement) as \emph{a priori} predictors of retention (RQ5), and
the synthetic-input variant of \method{Refresh-D}
\citep{wang2024translora} for when even task inputs are unavailable.

\textbf{Limitations.} Specialists are capped at 8B by the single-GPU
constraint and are adapted only with QLoRA at rank 16; full-parameter
fine-tuning perturbs weights at a much higher effective rank
\citep{biderman2024lora} and may well have a different transfer profile,
which we cannot speak to. Tasks are English, two per class, with the second-task suite measured
over three generations rather than four. The continuity criterion
rests on two fresh-pretraining hops (one inferred from release
documentation, one verified from a documented lineage), two
continuation hops, and one anneal-plus-soup hop; we report it as a
strongly supported, twice-replicated dichotomy with one measured
boundary, not yet a universal rate. The prompt-side comparison uses a
deliberately small search budget (six candidates, one task, one hop) and
selects on a dev slice, so it bounds the prompt asset from below rather
than characterizing prompt engineering in general. The amortized
decision model excludes re-validation and compliance costs.
\method{Refresh-D} is measured in its relabel variant, which assumes
task inputs were retained; this is the common but not the universal
enterprise case. UpgradeBench measures the tax and proposes no new
transfer method.

% =====================================================================
\section{Conclusion}
\label{sec:conclusion}

We introduced UpgradeBench and measured the base-model upgrade tax
along a complete fourteen-month open-weight release lineage, at two
model scales, across six enterprise tasks. The tax is a task property:
retrained reference scores rise only where the task--base coupling is
high and stay flat---within training noise---where it is low; an
apparent decline on CLINC150 did not survive three seeds. Specialization half-lives are long, bracketed to two to
three generations on the most base-bound task at 7--8B and observed
crossing zero at 1.5--1.8B, while a classification specialist keeps a
28.6-point lead after fourteen months of base progress. What licenses
adapter porting is pretraining continuity rather than architectural
compatibility: copying fails across independent pretraining runs in
all eleven task/scale cells in two families and retains 0.82--1.45
between checkpoints joined by continued pretraining in both families.
A prospectively specified distance ablation on OLMo's documented
training trajectory confirmed both regimes and located the boundary
between them: retention falls from 0.88--0.99 at a 46B-token
continuation to the no-adapter floor at 2.9T tokens, with the
subsequent anneal and model-soup merge adding no further damage;
portability is a distance budget, not a family property. Where copying is
unavailable, annotation-free (input-retaining) refresh reaches parity with
gold-label retraining at 39--331 consumer-GPU minutes; and the
prompt-side asset is an order of magnitude smaller than the weight
asset without being more durable. All code, adapters, per-query
records, CIs, and energy logs are released; the study cost 148.7
training GPU-hours (47.8\,kWh measured) plus 79.4 evaluation GPU-hours
on one RTX~4090.

% =====================================================================
\subsubsection*{Reproducibility Statement}
All experiments run on a single RTX~4090 with pinned software versions
(\S\ref{sec:setup}). We release training/evaluation scripts, all 98
adapter checkpoints, per-query prediction records and bootstrap CIs
for all 193 evaluation cells, dataset split seeds and manifests, the
three relabeled
Refresh-D corpora, the memorization-probe implementation and outputs,
and raw cost logs (wall-clock, peak VRAM, integrated energy) behind
every figure. Code and artifacts accompany the submission and will be
released publicly upon publication.

\subsubsection*{Ethics Statement}
All datasets are public under their stated licenses; no private or
personal data is used. The benchmark informs build-vs-buy decisions
with labor implications; we report costs transparently to support
informed choices. Fine-tuning can erode safety alignment
\citep{qi2024finetuning}; released adapters are task-scoped and carry
no safety claims. Total measured training energy: 47.8\,kWh.
\emph{Independence.} The study uses only publicly released checkpoints and public documentation; no internal training records or unreleased evaluations were used. Model and task selection criteria were fixed before the first experiment, and the OLMo panel provides an external, fully documented validation family. The first author is employed by Alibaba Group, whose Qwen models form the primary release series studied here; the employers had no role in the design, execution, analysis, or conclusions of this work.

\bibliographystyle{plainnat}
\bibliography{references}

% =====================================================================
\appendix

\section{Lineage details}
\label{app:lineages}
Architecture parameters in Table~\ref{tab:lineages} are read directly
from each checkpoint's \texttt{config.json} and are reproduced in the
release. Notes: (i)~the vocabulary size alternates
(151{,}936 $\to$ 152{,}064 $\to$ 152{,}064 $\to$ 151{,}936), so
embedding-row alignment differs by hop even though the tokenizer
family is constant; (ii)~Qwen1.5-7B uses full multi-head attention
(32 query, 32 key/value heads), making its KV cache $8\times$ larger
than Qwen2/2.5's GQA configuration, a deployment-relevant regression
that per-parameter comparisons miss, and in our runs the dominant
cost factor for long-prompt evaluation on a consumer GPU;
(iii)~Qwen3-8B introduces an explicit \texttt{head\_dim} of 128 and a
narrower FFN (12{,}288) than Qwen2.5 (18{,}944) despite being wider
overall; (iv)~Qwen2.5-7B-Instruct-1M (2025-01) matches
Qwen2.5-7B-Instruct in every architectural dimension and extends its
context window via continued pretraining, making it the continuation
release used in \S\ref{sec:results:copy}; (v)~the small-scale track
uses Qwen1.5-1.8B-Chat (MHA 16/16, hidden 2048, 24 layers),
Qwen2-1.5B-Instruct and Qwen2.5-1.5B-Instruct (identical: GQA 12/2,
hidden 1536, 28 layers, FFN 8960), and Qwen3-1.7B (GQA 16/8, hidden
2048, 28 layers).

\paragraph{Small-scale lineage scores.} Zero-shot / specialist by
generation. Banking77: .1984/.9247, .2942/.9260, .3614/.9273,
.4247/.9208. Spider: .1625/.4642, .3859/.5725, .3569/.6141,
.5455/.6509. xLAM-FC: .1245/.7335, .4800/.7685, .5880/.7940,
.6840/.8180. Copying on the small 2$\to$2.5 hop: Banking77 .0218
($R{=}-0.600$), Spider .3636 ($R{=}0.026$), xLAM .6075 ($R{=}0.095$).

\paragraph{OLMo validation lineage.} From \texttt{config.json}:
OLMo-1-7B and OLMo-1.7-7B are architecturally identical (hidden 4096,
32 layers, MHA 32/32, FFN 11008, vocabulary 50304); the OLMo-1.7 model
card states it was trained from scratch. OLMo-2-7B and its published
stage-1 checkpoints \texttt{stage1-step237000} and
\texttt{stage1-step248000} share one shape (hidden 4096, 32 layers,
MHA 32/32, FFN 11008, vocabulary 100352); the two stage-1 snapshots
are ${\sim}46$B training tokens apart within the same run, and the
final base adds the remaining stage-1 tokens, a stage-2 anneal on a
different data mix, and a model-soup merge \citep{olmo2025olmo2}. All
five are base checkpoints; training and evaluation use one fixed
plain-text format (system text, blank line, input, ``\texttt{Answer:}'')
with completion-only loss and the intent-task recipe of
\S\ref{sec:setup}. Zero-shot floor / specialist reference per checkpoint,
Banking77 then CLINC150 (lenient): OLMo-1 .0334/.8795, .1416/.8336;
OLMo-1.7 .2464/.9292, .2482/.9151; s1-237k .0045/.9279, .0091/.9062;
s1-248k .0097/.9305, .0120/.9078; s1-928646 .1432/.9289, .0315/.9204;
stage2-ingredient3 .3055/.9315, .2822/.9122; OLMo-2 main .3523/.9256,
.2575/.9125. The near-zero floors of mid-run checkpoints against the
substantial floors of the released models (OLMo-1.7, OLMo-2 main)
reflect instruction-following ability arriving late in pretraining;
retention is computed against each target's own floor and reference, so
the comparisons are unaffected.

\section{Task, prompt, and evaluator details}
\label{app:datasets}
Full prompt templates, label normalization, SQL execution comparator
(result-multiset equality, $10^{-4}$ float rounding, 10\,s timeout),
and call-set matcher (name equality; type-coerced, order-insensitive
argument comparison) as released. Banking77 5-shot exemplars are fixed
across all models (seed 42, five distinct labels). CLINC150 uses the
\texttt{plus} configuration (151 labels including out-of-scope);
lenient morphological matching changes the Qwen2.5 zero-shot floor
from .7173 (strict) to .7249 (lenient) and both metrics are released
per cell; under constrained label decoding the same floor is .7207
(\S\ref{sec:results:threats}). FinQA programs are normalized before exact match (whitespace,
trailing commas, numeric canonicalization); its adoption floors are
volatile across generations (.221/.093/.215) because zero-shot models
drift in and out of the dataset's DSL conventions, which we flag
wherever a FinQA ratio depends on the floor. glaive-FC is parsed from
the original corpus (arguments are single-quoted JSON inside the
assistant turn), with a 75/25 call/no-call mix; specialists and
baselines are graded on both emitting correct calls and refusing
correctly. An audit of our original random split found that 21.0\%
of test conversations (420/2{,}000) also appeared verbatim in
training---the corpus reuses templates and tool schemas heavily---so
all glaive-FC numbers in this paper use a re-split that groups
conversations by (masked user template, called function) and assigns
whole groups to train/val/test (7{,}580/505/2{,}000 from a
10{,}085-item exact-deduplicated pool, 5{,}428 groups). The re-split
has zero cross-split leakage at exact, normalized-exact, and group
level; the split manifest with SHA-256 hashes and the audit script are
released. Scores on the clean split are 0.6--5.4pp lower than on the
leaked split; the qualitative conclusion (a flat reference score
across generations) is unchanged. Overlong-completion drops: \S\ref{sec:cost}; additionally
CLINC150 0/10{,}000, FinQA 41/6{,}251, glaive-FC 96/10{,}000.

\section{Refresh-D details}
\label{app:refresh}
Relabel variant: teacher = the Qwen2.5-7B specialist for each task;
teacher inference over retained train+val inputs (5.5 / 112.6 / 152.5
minutes for Banking77 / Spider / xLAM); student recipe identical to
\S\ref{sec:setup}. Teacher--gold agreement: 97.45\%/92.6\%
(Banking77 train/val), 88.3\%/66.8\% (Spider exact-string; semantic
agreement is higher since equivalent SQL strings differ), 92.8\%/83.4\%
(xLAM). Three-way student--teacher--gold records released.

\section{Prompt-side pilot}
\label{sec:results:prompt}

On the Banking77 2.5$\to$3 hop we compare the weight asset against a
prompt asset (best of six system prompts selected on a 200-item dev
slice). The prompt asset is small and unstable: dev-selected prompts
gain at most +0.4pp over the default instruction on the new base
(.6474 vs.\ .6438 zero-shot), porting the old best prompt recovers
.6416, and the identity of the best prompt \emph{changes} across the
hop. The weight asset delivers +27--29pp over the same floors and
survives three generations (\S\ref{sec:results:halflife}). Both routes
pay an upgrade tax; on this task the weight asset is 30--50$\times$
larger in delivered advantage and no more fragile. (Scope: one task,
one hop, a deliberately small prompt-search budget; agentic harnesses
with tool scaffolds may behave differently, an extension the released
harness supports.)

\section{Memorization probe}
\label{app:contamination}
For 200 test items per task, each base receives the first 60\% of the
raw item text (no chat template) and greedily continues 24 tokens; we
report the longest-common-prefix ratio against the true continuation
and an 8-gram hit rate. Verbatim rates (LCP $>$ 0.5): Banking77
0--1\%, Spider 0\%, xLAM 3--5\% (with 8-gram rates 0.32--0.46 driven
by templated JSON tool syntax). Rates are approximately constant
across generations, so cross-generation comparisons are not
differentially contaminated. This probe detects verbatim memorization
only; it cannot rule out paraphrased exposure.

\section{Full run inventory and cost log}
\label{app:costlog}
98 training runs, 148.7 training GPU-hours (47.8\,kWh measured),
79.4 evaluation GPU-hours (energy not separately metered; the power
integral covers training only), 4.6 teacher-relabel GPU-hours; per-run
wall-clock, peak VRAM, and energy are tabulated in the released
\texttt{train\_log.json} files (longest single run:
glaive-FC/Qwen3-8B, 422.5\,min; largest energy: FinQA/Qwen3-8B,
1{,}255\,Wh; largest peak VRAM: FinQA/Qwen3-8B, 21.1\,GB; smallest
run: Spider ladder-256, 8.0\,min, 49.5\,Wh). 193 evaluation cells, 1.2--31.7 minutes each
under normal memory conditions; per-cell generation-token counts
logged. Three operational notes for single-GPU replication.
(1)~Exceeding physical VRAM triggers silent host-memory spill with
10--40$\times$ slowdown; with MHA checkpoints (KV cache $8\times$
GQA's) and ${\sim}$1{,}000-token prompts this is systematic at batch
32 on 24\,GB---the OLMo zero-shot floors cost 40.8 of the study's
GPU-hours this way before an automatic batch-shrink guard was added;
wall-clock is reported as measured. (2)~MSYS-based
shells exhaust fork resources after roughly a day of continuous
subprocess churn (the idempotent runner resumes cleanly). (3)~With
plain-text prompts, completion-only masking must tokenize prompt and
completion \emph{separately}: BPE merges a trailing-space prompt
boundary into the completion's first token, which silently masks that
token from the loss---training converges normally while the specialist
learns answers with their first token missing (reference self-evaluation
collapsed to 5--17\%). The wave-4 trainings were quarantined and rerun
after this was caught by the reference check; the released runner now
builds each example by concatenating separately tokenized prompt and
completion, which makes the training context a token-exact prefix of
the evaluation prompt by construction. One
evaluator bug (a type guard in the xLAM call matcher) was fixed and
the affected cell rerun; the changelog documents it.

\section{Per-cell confidence intervals}
\label{app:pertask}
Bootstrap 95\% CIs (10{,}000 resamples) for all 193 evaluation cells
are released as \texttt{ci\_table.json} (generated from the study
registry, which also lists the diagnostic cells excluded from the
study); representative half-widths:
Banking77 $\pm$0.9pp, Spider $\pm$2.8pp, xLAM $\pm$1.6pp. Seed
variance (3 seeds, Banking77/Qwen2.5): mean .9301, $\sigma$ = 0.12pp.

\section{Dataset licenses}
\label{app:licenses}
Banking77 (CC-BY-4.0); Spider (CC BY-SA 4.0); xLAM-function-calling-60k
(CC-BY-4.0, gated access via Hugging Face). License texts checked
2026-07; re-verify before redistribution.

\end{document}